\documentclass[sigconf]{acmart}
\usepackage{booktabs,wrapfig,multirow,colortbl,flushend}
\newcolumntype{C}[1]{>{\centering\arraybackslash}p{#1}}

\newcommand{\AM}[1]{\textcolor{green}{#1}}
\newcommand{\HH}{\mathbb{H}}
\newcommand{\RR}{\mathbb{R}}

\newcommand{\ww}{\boldsymbol{w}}
\newcommand{\vv}{\boldsymbol{v}}

\newcommand{\ee}{\boldsymbol{e}}
\newcommand{\xx}{\boldsymbol{x}}
\newcommand{\pphi}{\boldsymbol{\phi}}
\newcommand{\vvarphi}{\boldsymbol{\varphi}}
\acmConference[]{}
\acmYear{}
\copyrightyear{}
\begin{document}

\title{Taking Advantage of Multitask Learning \\ for Fair Classification}
%ness}
%\title{Enhancing Multitask Learning with Fairness Requirements}
%er Models via Multitask Learning}
%\title{Better and Fairer Models via Multitask Learning}
%\titlenote{Produces the permission block, and copyright information}
%\subtitle{Extended Abstract}
%\subtitlenote{The full version of the author's guide is available as \texttt{acmart.pdf} document}
\author{Luca Oneto}
%\authornote{Dr.~Trovato insisted his name be first.}
\affiliation{%
  \institution{University of Genoa}
%  \streetaddress{Via Opera Pia 11a}
  \city{Genoa, }
  \state{Italy}
  \postcode{16145}
}
\email{luca.oneto@unige.it}
\author{Michele Donini}
\affiliation{%
  \institution{Istituto Italiano di Tecnologia}
%  \streetaddress{Via Morego, 30}
  \city{Genoa, }
  \state{Italy}
  \postcode{16163}
}
\email{michele.donini@iit.it}
\author{Amon Elders}
\affiliation{%
  \institution{Istituto Italiano di Tecnologia}
%  \streetaddress{Via Morego, 30}
  \city{Genoa, }
  \state{Italy}
  \postcode{16163}
}
\email{amonelders@gmail.com}
\author{Massimiliano Pontil}
\affiliation{%
  \institution{Istituto Italiano di Tecnologia}
  \streetaddress{Via Morego, 30}
  \city{Genoa, }
  \state{Italy \\and}
  \postcode{16163}
}
\affiliation{%
\institution{University College London}
   \city{London, }
  \state{UK}
}
\email{massimiliano.pontil@iit.it}

\renewcommand{\shortauthors}{L. Oneto et al.}

\begin{abstract}
A central goal of algorithmic fairness is to reduce bias in automated decision making. 
An unavoidable tension exists between accuracy gains obtained by using sensitive information (e.g., gender or ethnic group) as part of a statistical model, and any commitment to protect these characteristics. 
Often, due to biases present in the data, using the sensitive information in the functional form of a classifier improves classification accuracy.
In this paper we show how it is possible to get the best of both worlds: optimize model accuracy and fairness without explicitly using the sensitive feature in the functional form of the model, thereby treating different individuals equally. 
Our method is based on two key ideas. On the one hand, we propose to use Multitask Learning (MTL), enhanced with fairness constraints, to jointly learn group specific classifiers that leverage information between sensitive groups. 
On the other hand, since learning group specific models might not be permitted, we propose to first predict the sensitive features by any learning method and then to use the predicted sensitive feature to train MTL with fairness constraints. 
This enables us to tackle fairness with a three-pronged approach, that is, by increasing accuracy on each group, enforcing measures of fairness during training, and protecting sensitive information during testing. Experimental results {on two real datasets} support our proposal, showing substantial improvements in both accuracy and fairness.
%We dub our approach Fair-MTL. Fair-MTL enables us to tackle fairness with a three-pronged approach, that is, by increasing accuracy on each group separately, by enforcing measures of fairness during training, as well as by treating different individuals equally. Experimental results support our proposal.
\end{abstract}
%\begin{CCSXML}
%<ccs2012>
%<concept>
%<concept_id>10010147.10010257.10010258.10010259.10010263</concept_id>
%<concept_desc>Computing methodologies~Supervised learning by classification</concept_desc>
%<concept_significance>500</concept_significance>
%</concept>
%<concept>
%<concept_id>10010147.10010257.10010258.10010262</concept_id>
%<concept_desc>Computing methodologies~Multi-task learning</concept_desc>
%<concept_significance>500</concept_significance>
%</concept>
%<concept>
%<concept_id>10010147.10010257.10010293.10010075</concept_id>
%<concept_desc>Computing methodologies~Kernel methods</concept_desc>
%<concept_significance>500</concept_significance>
%</concept>
%</ccs2012>
%\end{CCSXML}
%
%\ccsdesc[500]{Computing methodologies~Supervised learning by classification}
%\ccsdesc[500]{Computing methodologies~Multi-task learning}
%\ccsdesc[500]{Computing methodologies~Kernel methods}
%
%\keywords{Classification, Fairness, Sensitive Feature, Multitask Learning}
%
\maketitle
\section{Introduction}
\label{sec:intro}
%\vspace{-.15cm}
%
In recent years there has been a lot of interest in the problem of enhancing learning methods with ``fairness'' requirements, 
%imposing ``fairness'' constraints in learning methods, 
see~\cite{pleiss2017fairness,beutel2017data,hardt2016equality,feldman2015certifying,agarwal2017reductions,agarwal2018reductions,woodworth2017learning,zafar2017fairness,menon2018cost,zafar2017parity,bechavod2018Penalizing,zafar2017fairnessARXIV,kamishima2011fairness,kearns2017preventing,Prez-Suay2017Fair,dwork2018decoupled,berk2017convex,alabi2018optimizing,adebayo2016iterative,calmon2017optimized,kamiran2009classifying,zemel2013learning,kamiran2012data,kamiran2010classification} and references therein.
The general aim is to 
%enhance classifiers learned from data with fairness requirements, namely 
ensure that sensitive information (e.g.~knowledge about gender or ethnic group of an individual) does not ``unfairly'' 
influence the outcome of a learning algorithm. For example, if the learning problem is to predict what salary 
a person should earn based on her skills and previous employment records, we would like to build 
a model which does not unfairly use additional sensitive information such as gender or race.

A central question 
%key question \MP{which we aim to address in this paper} 
is how sensitive information should be used during the training and testing phases of a model. From a statistical perspective, sensitive information can improve model performance: removing this information may result in a less accurate model, without necessarily improving the fairness of the solution,~\cite{dwork2018decoupled,zafar2017fairness,pedreshi2008discrimination}.
%: simply removing this information may result in a less accurate model, without necessarily improving the fairness of the solution~\cite{dwork2018decoupled,zafar2017fairness,pedreshi2008discrimination}.
%From a legal point of view, 
However, it is well known, that in some jurisdictions using different classifiers, 
either explicitly or implicitly, for members of different groups, may not be permitted, we refer to the remark at page 3 in ~\cite{dwork2018decoupled} and references therein. 
%However, it is well known that, from a statistical perspective, using sensitive information can improve both accuracy and fairness measures~\cite{dwork2018decoupled,zafar2017fairness,pedreshi2008discrimination}. Our principal objective is then: can we develop a highly accurate machine learning method which exploits sensitive information while still treating different individuals equally? 
These imply that we can access the sensitive information during the training phase of a model but not during the testing phase. Our principal objective is then to optimize model accuracy while still protecting sensitive information in the data. 

As a first step towards not discriminating minority groups we focus on maximizing average accuracy with 
respect to each group as opposed to maximizing the overall accuracy~\cite{chouldechova2017fair}.
%%%
For the underlying generic learning method, we consider both Single Task Learning (STL) and Independent Task Learning (ITL).
While the latter independently learns a different function for each group, the former aims to learn a function that is common between all groups.
A well-known weakness of these methods is that they tend to generalize poorly on smaller groups: while STL {may  learn}
%often learns a shared 
a model which better represents the largest group, ITL may overfit minority groups 
%due to small sample size
~\cite{baxter2000model}. 
%\MP{<Do we verify this experimentally?>}
%%%
A common approach to overcome such limitations is 
offered by Multitask Learning (MTL), see~\cite{baxter2000model,caruana1997multitask,evgeniou2004regularized,bakker2003task,argyriou2008convex} and references therein.
This methodology leverages information between the groups (tasks) to learn more accurate models. 
Surprisingly, to the best of our knowledge, MTL has received little attention in the {algorithmic} fairness domain. 
We are only aware of the work~\cite{dwork2018decoupled} which proposes to learn different classifiers per group, combined with MTL to ameliorate the issue of potentially having too little data on minority groups. 
%\AM{However, a limitation of their method is that they use a different classifier for each group, thereby not treating different individuals equally.}
%which proposes a decoupled approach to learn different classifiers, combined with MTL to ameliorate the issue of potentially having too little data on minority groups. Then, to retrieve fair classifiers, a joint loss function encoding a notion of fairness is minimized over this set of classifiers. 

%In this paper 
We build upon a particular instance of MTL which jointly learns a shared model between 
the groups as well as a specific model per group. We show how fairness constraints, measured with Equalized Odds or Equal Opportunities introduced in~\cite{hardt2016equality}, can be built in MTL directly during the training phase. 
This is in contrast to other approaches which impose the fairness constraint as a post-processing step~\cite{pleiss2017fairness,beutel2017data,hardt2016equality,feldman2015certifying} 
or by modifying the data representation before employing standard machine learning methods~\cite{adebayo2016iterative,calmon2017optimized,kamiran2009classifying,zemel2013learning,kamiran2012data,kamiran2010classification}. 
In many recent works~\cite{donini2018empirical,agarwal2017reductions,agarwal2018reductions,woodworth2017learning,zafar2017fairness,menon2018cost,zafar2017parity,bechavod2018Penalizing,zafar2017fairnessARXIV,kamishima2011fairness,kearns2017preventing,Prez-Suay2017Fair,dwork2018decoupled,berk2017convex,alabi2018optimizing,dwork2018decoupled} it has been shown how to enforce these constraints during the learning phase of a classifier. Here we opt for the approach proposed in~\cite{donini2018empirical} since it is convex, theoretically grounded, and performs favorably against state-of-the-art alternatives. {We present experiments on two real-datasets} which demonstrate that the shared classifier learned by MTL works better than STL and in turn MTL's group specific classifiers perform better than both ITL as well as the shared MTL model. 
{These results are in line with previous studies on MTL, which suggest the benefit offered by this methodology, see~\cite{Efros,evgeniou2004regularized,donini2016distributed} and references therein. Moreover, we observe that the fairness constraint is effective in controlling the fairness measure.}

Unfortunately, as remarked before, all the models which employ the sensitive feature in the testing phase may not be adoptable. 
Independent models cannot be employed since we are using different classifiers for members of different groups.
Even the shared model may not be a feasible option, if the sensitive feature is used as a predictor
(e.g. if the model is linear, including the sensitive feature entails using a group specific threshold). Therefore, the only feasible\footnote{The sensitive feature may not be available in the testing phase or it might not be possible to use it as a predictor in the model due to legal requirements~\cite{dwork2018decoupled}.}
option would be to learn a shared model based on the non-sensitive features.
This constraint may limit our ability to learn classifiers of high generalization ability.
\begin{figure}[h]
\begin{center}
\includegraphics[width=.9\columnwidth]{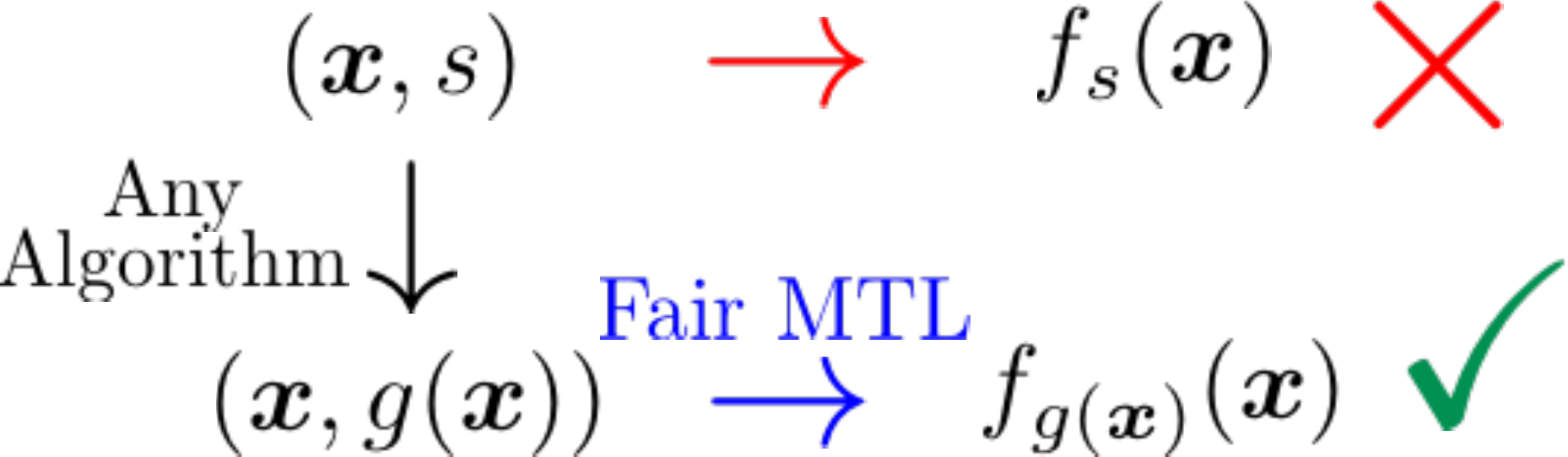}
\end{center}
\caption{Our proposal in a graphical abstract: rather than using the sensitive feature $s$ as a predictor we propose to learn, with any learning algorithm, a function $g$, which captures the relationship between $\xx$ and $s$, and then use $g(\xx)$, instead of $s$, to learn group specific models via MTL.}
\label{fig:GraphicalAbstract}
\end{figure}
In order to overcome such limitations, we propose to first use the non-sensitive features to predict the value of the sensitive one and then use the predicted sensitive feature to learn group specific models via MTL. 
The proposal is depicted in the graphical abstract of Figure~\ref{fig:GraphicalAbstract}.
We experimentally demonstrate that the proposed approach matches the classification accuracy of the best performing model which uses the sensitive information during testing, in addition to further improving upon measures of fairness. %\AM{<add few lines on explanation/insights on why this works/what made us think of this method?>}
%\AM{We experimentally demonstrate that the proposed approach favorably performs against STL, ITL, improving upon both accuracy, and measures of fairness. Perhaps even more surprising, we also find that the proposed approach substantially improves upon the fairness measures of the best performing, but potentially non-permitted MTL model, namely, the method that does use the true sensitive feature as part of the model, while at the same time matching classification accuracy. We argue that, on the one hand, in the case of high accuracy, our method is able to enforce measures of fairness, as well as share information between groups. On the other hand, in the case of low accuracy, our method effectively performs a randomization procedure, making the model essentially blind with respect to the sensitive features, further improving upon the Equal Opportunity and Equal Odds fairness measures.}

The rest of the paper is organized as follows.
Section~\ref{sec:Preliminaries} presents some preliminary definitions and notions concerning the fair classification framework.
Section~\ref{sec:Problem} outlines the central problem that we face in the paper: exploiting the sensitive feature while still treating different groups equally.
Section~\ref{sec:Methodology} presents our proposal: predicting the sensitive feature based on the non-sensitive ones and then exploiting MTL with fairness constraints in order to increase both accuracy and fairness measures (see Figure~\ref{fig:GraphicalAbstract}).
In Section~\ref{sec:Experimental} we test the proposal on two well known fairness related datasets (Adult and COMPAS) demonstrating the potentiality of it. 
We conclude the paper with a brief discussion in Section~\ref{sec:Discussion}.

%In summary our main contributions are to show how 
%
\section{Preliminaries}
\label{sec:Preliminaries}
We let $\mathcal{D} {=} \{ (\xx_1,s_1,y_1),$ $\dots,$ $(\boldsymbol{x}_n,s_n,y_n) \}$ be a training set formed by $n$ samples drawn independently from an unknown probability distribution $\mu$ over $\mathcal{X} {\times} \mathcal{S} {\times} \mathcal{Y}$, where $\mathcal{Y} {=} \{ -1, +1 \}$ is the set of binary output labels, $\mathcal{S} {=} \{1, \dots , k \}$ represents group membership, and $\mathcal{X}$ is the input space.
%For every $t {\in} {\cal S}$ and operator $ {\diamond} {\in} \{-,+\}$, we define the subset of training points with sensitive feature equal to $t$ as $\mathcal{D}_{t} {=} \{(\boldsymbol{x},s,y) : (\boldsymbol{x},s,y) {\in} \mathcal{D}, s {=} t \}$ with $n_{t} {=} |\mathcal{D}_{t}|$ and the subset of training point negatively and positively labeled with sensitive feature equal to $t$ as $\mathcal{D}^\diamond_t {=} \{(\boldsymbol{x},s,y) : (\boldsymbol{x},s,y) {\in} \mathcal{D}, s {=} t, y {=} {\diamond} 1 \}$ likewise with $n^\diamond_t {=} |\mathcal{D}^\diamond_t|$ .

For every $t {\in} {\mathcal{S}}$ and operator $ {\diamond} {\in} \{-,+\}$, we define the subset of training points with sensitive feature equal to $t$ as $\mathcal{D}_{t} {=} \{(\boldsymbol{x},s,y) : (\boldsymbol{x},s,y) {\in} \mathcal{D}, s {=} t \}$ and the subset of training point negatively and positively labeled with sensitive feature equal to $t$ as $\mathcal{D}^\diamond_t {=} \{(\boldsymbol{x},s,y) : (\boldsymbol{x},s,y) {\in} \mathcal{D}, s {=} t, y {=} {\diamond} 1 \}$. We also let $n_{t} {=} |\mathcal{D}_{t}|$ and for ${\diamond} {\in} \{-,+\}$, we let $n^\diamond_t {=} |\mathcal{D}^\diamond_t|$ .

Let us consider a function (or model) $f: \mathcal{X} {\times} \mathcal{S} {\rightarrow} \mathbb{R}$ chosen from a set $\mathcal{F}$ of possible models. The error (risk) of $f$ is measured by a prescribed loss function $\ell:\mathbb{R} {\times} \mathcal{Y} {\rightarrow} \mathbb{R}$ .
The average accuracy with respect to each group of a model $L(f)$, together with its empirical counterparts $\widehat{L}(f)$, are defined respectively as
\begin{align}
&  L(f) {=} \frac{1}{k} \sum_{t \in \mathcal{S}} L_t(f), \quad 
L_t(f) = \mathbb{E} \left[\ell(f(\xx,s,y)\ | \ s=t \right], \ t {\in} \mathcal{S}, \nonumber \end{align}
and
\begin{align}
& \widehat{L}(f) {=} \frac{1}{k} \sum_{t \in \mathcal{S}} \widehat{L}_t(f), \quad 
\widehat{L}_t(f) {=} \frac{1}{n_t} \sum_{(\xx,s,y) \in \mathcal{D}_t} \ell(f(\boldsymbol{x},s),y), \ t {\in} \mathcal{S}.
\nonumber
\end{align}
The fairness of the model can be measured w.r.t. many notions of fairness as mentioned in Section~\ref{sec:intro}.
In this work we choose to opt for the Equal Opportunity (EOp) and the Equal Odds (EOd).
For ${\diamond} \in \{-,+\}$, the EOp$^{\diamond}$ constraint is defined as~\cite{hardt2016equality}
\begin{align}\label{eq:1}
\mathbb{P}\{f(\boldsymbol{x},s) {>} 0\ | \ s {=} 1, y{=} {\diamond} 1 \}{ =} \dots {=} \mathbb{P}\{f(\boldsymbol{x},s) {>} 0\ | \ s {=} k, y{=} {\diamond} 1 \},
\end{align}
where $ {\diamond} \in \{-,+\}$.  
%\MP{I would remove until end of this:} since we can define the EOp of the positively (EOp$^+$) or negatively (EOp$^-$) labeled samples. 
The EOd, instead is just the 
%contemporary 
concurrent verification of the EOp$^+$ and EOp$^-$, then $\forall {\diamond} {\in} \{-,+\}$
\begin{align}\label{eq:2}
\mathbb{P}\{f(\boldsymbol{x},s) {>} 0\ | \ s {=} 1, y{=} {\diamond} 1 \} {=} \dots {=} \mathbb{P}\{f(\boldsymbol{x},s) {>} 0\ | \ s {=} k, y{=} {\diamond} 1 \}.
\end{align}
Since a model $f$, in general, will not be able to exactly fulfill the EOp$^{+}$ with ${\diamond} \in \{-,+\}$ nor the EOd constraints we define the Difference of EOp$^ {\diamond}$ (DEOp$^{\diamond}$) with $ {\diamond} \in \{-,+\}$ as
\begin{align}
{\rm DEOp}^{\diamond} = \sum_{t{\in} \mathcal{S}} \left| \mathbb{P}\{
{\hat y = y }
%f(\boldsymbol{x},s) {>} 0 
| s {=} t, y{=} {\diamond} 1 \} {-} 
 \frac{1}{|\mathcal{S}|} \sum_{t' {\in} \mathcal{S}} \mathbb{P}\{
{\hat y = y }
%f(\boldsymbol{x},s) {>} 0 
| s {=} t', y{=} {\diamond} 1 \} \right|,  \nonumber
\end{align}
where ${\hat y} = {\rm sign}(f(\boldsymbol{x},s))$. Finally, the Difference of EOd (DEOd) is defined as 
\begin{align*}
{\rm DEOd} = \frac{{\rm DEOp}^+ + {\rm DEOp}^-}{2}. 
\end{align*}
\section{Paradigm}
%\AM{Instead of calling the section "problem", call it the solution, then it becomes more clear what our result is? "treating different individuals equally/how to treat different individuals equally"}
\label{sec:Problem}
A central problem, when learning a model $f$ from data under fairness requirements, 
is that using a different classification method, or even using different weights on attributes 
for members of different groups may not be allowed for certain classification tasks~\cite{dwork2018decoupled}.
%MP: Let's discuss
%\AM{Dwork leaves the question open if it's illegal and explicitly says that 'the question of whether or not to use decoupled classifiers is orthogonal to our work'. Need another paper perhaps? See e.g., 'Zafar: Fairness Constraints: Mechanisms for Fair Classification', they cite a paper on this issue by ''}.
In other words, it may not be permitted to use the sensitive feature explicitly or implicitly in the functional form of the 
model\footnote{Note that, for clarity, the above limitation is imposed only when making predictions with $f$.
During the training phase, the sensitive information can and should be used to guide the choice of model parameters.}. 
This means that $f$ {should be} a function of $\xx$ only, that is, $f(\xx,s) = f(\xx)$.

For instance, if $\mathcal{X}{=}\mathbb{R}^d$ and the sensitive feature is encoded with a one-hot encoding, and we use a linear classifier then
\begin{equation}
f(\xx,s) = \ww \cdot \xx + b_s, \quad \ww {\in} \mathbb{R}^d,\ b_s {\in} \mathbb{R},
\nonumber
\end{equation}
which is forbidden since the model involves a different bias for each of the sensitive groups. 
The problem is even more apparent when we use a different model per each group, namely we set
\begin{equation}
f(\xx,s) = {\ww}_s \cdot \xx + b_s, \quad \ww_s {\in} \mathbb{R}^d,\ b_s {\in} \mathbb{R}.
\label{eq:functionalMTLLuca}
\end{equation}
Unfortunately, the above requirement can be highly constraining, resulting in a model with poor accuracy. 
In practice, due to bias present in the data, learning a model 
which involves the sensitive feature in its functional form may substantially improve model accuracy.
%\AM{remove sentence, I think it's ill-defined what 'bias in the data' actually is, but if we remove it here, then have to remove it everywhere...}

Our proposal to overcome the above limitation
%, without encountering any legal issues,
is to use the input $\xx$ to predict the sensitive group $s$.
That is, we learn a function $g: \mathcal{X} {\rightarrow} \mathcal{S}$, such that $\hat{s} = g(\xx)$ is the prediction of the sensitive feature of $\xx$. Therefore, our method replaces the specific model $f(\xx,s)$ with the composite model $h(x) \equiv f(\xx,g(\xx))$, thereby treating different individuals equally. Indeed if $(x,t)$ and $(x',t')$ are two instances, then $h(x) \approx h(x')$ provided $x \approx x'$ irrespective of the values of $t$ and $t'$. Hence, we can freely use $\hat{s}$ in the functional 
%without encountering any legal issues
since, during the testing phase, we do not require any knowledge of $s$. As we shall see, on the one hand, in the regions of the input space where the classifier $g$ predicts well, 
this approach allows us to exploit MTL to learn group specific models.
On the other hand, when the prediction error is high, this approach acts as a randomization 
procedure\footnote{A random prediction $\hat{s}$ of $s$ is substituted in the functional 
form of Eq.~\eqref{eq:functionalMTLLuca} which then randomly selects one of the group specific models, 
transforming the function form in a randomized shared model.
Suppose we have many classifiers $f(\cdot, s)$ and a function $g$ which chooses which classifier to use.
If one assumes that $g(\boldsymbol{x})$ is purely random, then $f(\cdot,\boldsymbol{x})$ is a randomized classifier.
Therefore if $g$ has a high error rate, $g$ is unable to predict the sensitive feature.
Consequently $f(\cdot,\boldsymbol{x})$ is just a shared classifier composed of many functions chosen at random by $g$.} which, as we will empirically show, improves the fairness measure of the overall model.

%\MP{This phrase uses the terminology STL ITL and MTL introduced in the intro so maybe we can put it back in the introduction:}
%\MP{<REMOVE OR MOVE TO SEC 4?>
In this paper we investigate (i) the effect of having the sensitive feature as part of 
the functional form of the model, (ii) the effect of using a shared model between the groups 
or a different model per group, (iii) the effect of learning a shared model with STL or MTL 
and the effect of learning group specific models with ITL or MTL, and (iv) the effect of using the predicted sensitive 
feature instead of its actual value inside the functional form of the model. 
Then we will show that it is possible to take the best result of the different approaches 
with substantial benefits in terms of both model accuracy and fairness, while still 
treating different individuals equally.%}
%without unfairly treating members of different groups.
%compromising the legality of the solution. 
%We show that by combining the results obtained in (i-iv) it is possible to gain substantial 
%benefits both in terms of accuracy as well as fairness, without compromising legality.\AM{last two sentences are the same}
%\AM{Maybe a comment/intuition/story somewhere for why this might be the case?}
%MASSI{Let's discuss in skype tomorrow maybe}
%
%\vspace{-.15cm}
\section{Methodology}
\label{sec:Methodology}
%\vspace{-.15cm}
%
In this section, we describe our approach to learning fair and accurate models and highlight the connection to MTL~\cite{evgeniou2004regularized}. 
We consider the following functional form 
\begin{align}
f(\boldsymbol{x},s) = 
\ww \cdot \pphi(\xx,s), \quad (\xx,s) {\in} {\mathcal{X}} {\times} {\mathcal{S}},
\label{eq:funzioni}
\end{align}
where $``\cdot"$ is the inner product between two vectors in a Hilbert 
space\footnote{For all intents and purposes, one may also assume throughout that $\HH=\RR^d$, the standard $d$-dimensional vector space, 
for some positive integer $d$.} $\HH$, $\ww \in \HH$ is a vector of parameters, 
and $\pphi: {\mathcal{X}} \times {\mathcal S} \rightarrow \HH$ is a prescribed feature mapping\footnote{In practice, a bias term (threshold) can be added to $f(\xx,s)$ (which may depend on $s$) but to ease our presentation we do not include it if not necessary.}.

We can then learn the parameter vector $\ww$ by regularized empirical risk minimization, using the square Euclidean norm of the parameter vector $\| \ww \|^2$ as the regularizer.
The generality of this approach comes from the general form of the feature mapping $\pphi: \mathcal{X} {\times} \mathcal{S} {\rightarrow} \HH$ which may be implicitly defined by a kernel function, see e.g.~\cite{shawe2004kernel,smola2001} and references therein. In the following, first we will briefly discuss three approaches for learning the parameter vector which correspond to the three methods investigated in this paper.
Then, we will explain how these methods can be enhanced with fairness constraints. 
%
%\vspace{-.15cm}
\subsection{Single Task Learning}
\label{subsec:STL}
%\vspace{-.15cm}
%
As we argued above, we may not be allowed to explicitly use the sensitive feature in the functional form of the model. 
A simple approach to overcome this problem, would be to train a shared model between the groups, that is, we choose $\pphi(\xx,s) {=} \vvarphi(\xx)$ and $\ww {=} \ww_0$ in Eq.~\eqref{eq:funzioni}, where $\vvarphi: \mathcal{X} {\rightarrow} \HH$ and $\ww_0 {\in} \HH$, so that $f(\xx,s) {=} \ww_0 \cdot \vvarphi(\xx)$ (a potentially unregularized threshold may be built in the feature map to include a bias term).
We learn the model parameters by solving the Tikhonov regularization problem\footnote{With a little abuse of notation we replace in the risk definitions the function with its parameter vector.}
\begin{equation}
\label{eq:STL}
\min_{\ww_0 \in \HH} \quad \hat{L}(\ww_0) + \rho \| \ww_0 \|^2,
\end{equation}
where $\rho {\in} [0, \infty)$ is a regularization parameter.
This method, which we will call Single Task Learning (STL), searches for the linear separator which minimizes a trade-off between the empirical average risk per group and the complexity (smoothness) of the models.

As we shall see in our experiments below\AM{}, STL performs poorly, because it does not capture variance across groups. 
A slight variation which may improve performance is to introduce group specific thresholds. However, we remark again that this approach may not be permitted. Specifically, we choose $\pphi(\xx,s) {=} (\vvarphi(\xx),\ee_s)$ 
and $\ww {=} (\ww_0,\mathbf{b})$ where $\ee_1,\dots,\ee_S$ are the canonical basis vectors in $\mathbb{R}^k$ and $\mathbf{b} {=} (b_1, \dots, b_k) {\in} \mathbb{R}^k$, so that $f(\xx,s) = \ww_0 \cdot \vvarphi(\xx) + b_s$.
%\MP{remove this?:}\AM{I agree} Note that, as argued before, if we substitute $\hat{s}$ to $s$ in this last functional form.% then this model becomes legal.
%
%\vspace{-.15cm}
\subsection{Independent Task Learning}
\label{subsec:ITL}
%\vspace{-.15cm}
%
An approach to overcome the potentially underfitting performance of STL is to 
learn different models for each of the groups, we refer to this approach as independent task learning (ITL).
It corresponds to setting $\pphi(\xx,s) {=} ({\bf 0}_{s-1},\vvarphi(\xx),{\bf 0}_{k-s})$ and $\ww {=} 
(\ww_1, \dots, \ww_k)$ in Eq.~\eqref{eq:funzioni}, where $\vvarphi: \mathcal{X} {\rightarrow} \HH$ and $\ww_s {\in} \HH$ $\forall s {\in} \mathcal{S}$, so that $f(\xx,s) {=} \ww_s \cdot \vvarphi(\xx)$.
As before, the feature map may account for a constant component to accommodate a threshold for each of the groups. 
To find the vectors $\ww_s$ we solve $k$ independent Tikhonov regularization problems of the form 
\begin{equation}
\label{eq:ITL}
\min_{\ww_s \in \HH} \quad \hat {L}_s(\ww_s) + \rho \|\ww_s\|^2.
\end{equation}
Note that, similar to STL, if we substitute $\hat{s}$ to $s$ in this last functional form then the method treats members of different groups equally, since , as we mentioned before, learning independent models may not be allowed. Furthermore, we remark that from a statistical point of view, minority groups (small sample sizes) will be prone to overfitting. 
Nevertheless, as we shall see, ITL works better than STL in our experiments, suggesting that there is a lot of bias in the data. Still one would expect that by leveraging similarities between the groups ITL can be further improved. We discuss this next.
%
%\vspace{-.15cm}
\subsection{Multitask Learning}
\label{subsec:MTL}
%\vspace{-.15cm}
%
Let us now discuss the multitask learning approach used in the paper, which is based on regularization around a common mean~\cite{evgeniou2004regularized}.
We choose $\pphi(\xx,s) {=} (\vvarphi(\xx),{\bf 0}_{s-1},\vvarphi(\xx),{\bf 0}_{k-s})$ and $\ww {=} (\ww_0,\vv_1, \dots, \vv_k)$ in Eq.~\eqref{eq:funzioni}, where $\ww_0 {\in} \HH$ and $\vv_s {\in} \HH$ $\forall s {\in} \mathcal{S}$, so that $f(\xx,s) = \ww_0 \cdot \vvarphi(\xx) + \vv_t \cdot \vvarphi(\xx)$. 
MTL jointly learns a {\em shared} model $\ww_0$ as well as {\em task specific} models $\ww_s {=} \ww_0 {+} \vv_s {\in} \HH$ $\forall s {\in} \mathcal{S}$ by encouraging the specific models and the shared model to be close to each other. 
To this end, we solve the following Tikhonov regularization problem
\begin{align}
\label{eq:MTL}
\min_{\ww_0,\ww_1,\dots,\ww_S \in \HH} \ \ &  \theta \hat{L}(\ww_0) {+} (1{-}\theta) \frac{1}{k} 
\sum_{s =1}^k \hat{L}_s(\ww_s) \nonumber \\
& {+} \rho \left[\lambda \|\ww_0\|^2 {+} (1{-}\lambda) 
\frac{1} {k}
\sum_{s=1}^k \| \ww_s \|^2\right],
\end{align}
where the parameter $\lambda {\in} [0,1]$ forces the dependency between shared and specific models and the parameter $\theta {\in} [0,1]$ captures the relative importance of the loss of the shared model and the group-specific models.
This MTL approach is general enough to include STL and ITL, which are recovered by setting $\lambda{=}\theta{=}1$ and $\lambda {=} \theta {=} 0$, respectively.
Similar to STL and ITL, regularized group specific thresholds could be added in the shared model and in the group specific models.

Again, note that the group specific models trained by MTL may not be permitted. Likewise the shared model trained by MTL may not be permitted if we include
%\MP{<append or include?>}
the sensitive variable to the input. %\MP{<expand a bit this?:>} 
However if the sensitive variable is predicted from an external 
classifier and then MTL retrained with the predicted values, then this model treats different groups equally (see Figure~\ref{fig:GraphicalAbstract}).
%
%\vspace{-.15cm}
\subsection{Adding Fairness Constraints}
\label{subsec:EO}
%\vspace{-.15cm}
%
Note that both STL, ITL and MTL problems are convex provided the the loss function used to measure the empirical errors ${\hat L}$ and ${\hat L}_s$ in Eqns.~\eqref{eq:STL},~\eqref{eq:ITL},~and~\eqref{eq:MTL} are convex. 
Since we are dealing with binary classification problems, we will use the hinge loss (see e.g.~\cite{shalev2014understanding}), 
which is defined as $\ell(f(\boldsymbol{x},s),y) = \max\big(0,1{-}y f(\xx,s)\big)$.

In many recent papers~\cite{pleiss2017fairness,beutel2017data,hardt2016equality,feldman2015certifying,agarwal2017reductions,agarwal2018reductions,woodworth2017learning,zafar2017fairness,menon2018cost,zafar2017parity,bechavod2018Penalizing,zafar2017fairnessARXIV,kamishima2011fairness,kearns2017preventing,Prez-Suay2017Fair,dwork2018decoupled,berk2017convex,alabi2018optimizing,adebayo2016iterative,calmon2017optimized,kamiran2009classifying,zemel2013learning,kamiran2012data,kamiran2010classification,donini2018empirical} it has been shown how to enforce EOp$^{ {\diamond}}$ constraints for $ {\diamond} {\in} \{-,+\}$, during the learning phase of the model $f {\in} \mathcal{F}$.
Here we build upon the approach proposed in~\cite{donini2018empirical} since it is convex, theoretically grounded, and showed to perform favorably against state-of-the-art alternatives.
%The approach is quite simple but effective but as we will demonstrate empirically, effective.
To this end, we first observe that
\begin{align}\label{eq:3}
& \mathbb{P}\{f(\boldsymbol{x},s) > 0\ | \ s = t, y= {\diamond} 1 \} \nonumber \\
& = 1 {-} \mathbb{E} \left\{ \ell_h(f(\boldsymbol{x},s),y)\ | \ s=t, y= {\diamond} 1 \right\} \nonumber \\
& = 1 {-} L_t(f), \quad t {\in} \mathcal{S},
\end{align}
where $\ell_h(f(\boldsymbol{x},s),y) {=} [y f(\boldsymbol{x},s) {\leq} 0]$ is the hard loss function.
Then, by substituting Eq.~\eqref{eq:3} in Eqs.~\eqref{eq:1} and~\eqref{eq:2}, replacing the deterministic quantities with their empirical counterpart, and by approximating the hard loss function $\ell_h$ with the linear one $\ell_l {=} (1{-}y f(\boldsymbol{x},s))/2$ we have that the convex EOp$^ {\diamond}$ constraints with $ {\diamond} \in \{-,+\}$ is defined as follows
\begin{align}\label{eq:4}
\frac{1}{n^\diamond_1} \sum_{(\boldsymbol{x},s,y) \in \mathcal{D}^\diamond_1} f(\boldsymbol{x},s) = \dots = \frac{1}{n^\diamond_k} \sum_{(\boldsymbol{x},s,y) \in \mathcal{D}^\diamond_k} f(\boldsymbol{x},s),
%\AM{, \text{I changed it for the other as well}}
\end{align}
while for the EOd we just have to enforce both the EOp$^+$ and EOp$^-$ constraints.

In order to plug the constraint of 
Eq.~\eqref{eq:4} inside STL, ITL and MTL we first define
the quantities
\begin{align}
\boldsymbol{u}^\diamond_t = \frac{1}{n^\diamond_t} \sum_{ (\boldsymbol{x},s) \in \mathcal{D}^\diamond_t}
\boldsymbol{\varphi}(\boldsymbol{x}), \quad t {\in} \mathcal{S},\ \diamond {\in} \{-,+\}.
\end{align}
It is then straightforward to show that if we wish to enforce the EOp$^{\diamond}$ constraint 
onto the shared model one has to add these $(k{-}1)$ constraints to the STL and MTL
\begin{align}
\boldsymbol{w}_0 \cdot (\boldsymbol{u}^\diamond_1 - \boldsymbol{u}^\diamond_2 ) = 0
\ \wedge \ \dots \ \wedge \
\boldsymbol{w}_0 \cdot (\boldsymbol{u}^\diamond_1 - \boldsymbol{u}^\diamond_k ) = 0.
\end{align}
We remark again that for the EOd constraints we just have to insert $\text{EOp$^+$} \wedge \text{EOp$^-$}$ 
which means $2(k{-}1)$ constraints.

If, instead, we want to enforce the EOp$^ {\diamond}$ constraint onto group specific models 
we have to add these $(k{-}1)$ constraints to the MTL and ITL
\begin{align}
\boldsymbol{w}_1 \cdot \boldsymbol{u}^\diamond_1 = \boldsymbol{w}_2 \cdot \boldsymbol{u}^\diamond_2
\ \wedge \ \dots \ \wedge \
\boldsymbol{w}_1 \cdot \boldsymbol{u}^\diamond_1 = \boldsymbol{w}_k \cdot \boldsymbol{u}^\diamond_k,
\end{align}
while for the EOd we just have to insert $\text{EOp$^+$} \wedge \text{EOp$^-$}$.
%which means $2(k{-}1)$ constraints.

Al last we note that by the representer theorem, as shown in~\cite{donini2018empirical}, 
it is straightforward to derive the kernelized version of the fair STL, ITL, and MTL convex problems 
which can be solved with any solver, in our case CPLEX~\cite{cplex2018ibm}.
%
%\vspace{-.15cm}
\section{Experiments}
\label{sec:Experimental}
%\vspace{-.15cm}
%
% Do not repeat the list and refere to to Section 3 or Section 1
The aim of the experiments is to address the questions raised before.
Namely, we wish to: (a) study the effect of using the sensitive feature as a way to bias the decision of a common model or to learn group specific models, (b) show the advantage of training either the shared or group specific models via MTL, and
(c) show that MTL can be effectively used even when the sensitive feature is not available during testing by predicting the sensitive feature based on the non-sensitive ones. 

\subsection{Datasets and Setting}

%In all experiments, we performed a 10-fold cross validation (CV) to select the best hyperparameters\footnote{The ranges of hyperparameters used in the validation procedure of STL, MTL, and ITL are $\rho {\in} \{10^{-6.0}, 10^{-5.5}, \dots , 10^{+6.0} \}$ and $\lambda,\theta {\in} \{0, 2^{-15}, 2^{-14}, \dots , 2^{-1}, 1{-}2^{-2}, \dots , 1{-}2^{-15}, 1 \}$.}.

We employed the Adult dataset from the UCI repository\footnote{\url{https://archive.ics.uci.edu/ml/datasets/adult}} and the Correctional Offender Management Profilingfor Alternative Sanctions (COMPAS) dataset\footnote{\url{www.propublica.org/article/how-we-analyzed-the-compas-recidivism-algorithm}}.

The Adult dataset contains $14$ features concerning demographic characteristics of $45222$ instances ($32561$ for training and $12661$ for testing), $2$ features, Gender (G) and Race (R), can be considered sensitive.
The task is to predict if a person has an income per year that is more (or less) than $50,000\$$.
Some statistics of the adult dataset with reference to the sensitive features are reported in Table~\ref{tab:statistics}. 
\begin{table}
\centering
\setlength{\tabcolsep}{0.08cm}
\renewcommand{\arraystretch}{1}
\begin{tabular}{|c|l|r|c|c|}
\hline
\hline
Sens. & Group & $\%$ \\
\hline
\hline
\multirow{2}{*}{G} 
& Male (M) & 66.9 \\
& Female(F) & 33.2 \\
\hline
\multirow{5}{*}{R} 
& White (W) & 85.5 \\
& Black (B) & 9.6 \\
& Asian-Pac-Islander (API) & 3.1 \\
& Amer-Indian-Eskimo (AIE) & 1.0 \\
& Other (O) & 0.8 \\
\hline
& W\&M & 58.8 \\ 
& W\&F & 26.7 \\ 
& B\&M & 4.9 \\
& B\&F & 4.7 \\
G+R & API\&M & 2.1 \\
& API\&F & 1.1 \\ 
& AIE\&M & 0.6 \\
& AIE\&F & 0.4 \\
& O\&M & 0.5 \\
& O\&F & 0.3 \\
\hline
\hline
\end{tabular}
\caption{Adult dataset: statistics with reference to the sensitive features.}
\label{tab:statistics}
\end{table}

The COMPAS dataset is constructed by the commercial algorithm COMPAS, which is used by judges and parole officers for scoring criminal defendants likelihood of reoffending (recidivism).
It has been shown that the algorithm is biased in favor of white defendants based on a 2-years follow up study.
This dataset contains variables used by the COMPAS algorithm in scoring defendants, along with their outcomes within two years of the decision, for over 10000 criminal defendants in Broward County, Florida. 
In the original data, 3 subsets are provided.
We concentrate on the one that includes only violent recividism.
Table~\ref{tab:COMPAS:statistics}, analogously to Table~\ref{tab:statistics}, reports the statistics with reference to the sensitive features.
\begin{table}
\centering
\setlength{\tabcolsep}{0.08cm}
\renewcommand{\arraystretch}{1}
\begin{tabular}{|c|l|r|c|c|}
\hline
\hline
Sens. & Group & $\%$ \\
\hline
\hline
\multirow{2}{*}{G} 
& Female (F) & 19.34 \\
& Male (M) & 80.66 \\
\hline
\multirow{5}{*}{R} 
& African-American (AA)	 	& 51.23 	\\
& Asian (A) 				& 0.44 	\\
& Caucasian (C)			& 34.02	\\
& Hispanic (H)			& 8.83 	\\
& Native American (NA) 		& 0.25 	\\
& Other  (O)				& 5.23 	\\
\hline
& Female African-American & 9.04  \\
& Female Asian  & 0.03 \\
& Female Caucasian  & 7.86 \\
& Female Hispanic & 1.48 \\
& Female Native American & 0.06 \\
& Female Other & 0.93 \\
G+R & Male African-American & 42.20 \\
& Male Asian & 0.45 \\
& Male Caucasian & 26.16 \\
& Male Hispanic & 7.40 \\
& Male Native American & 0.19 \\
& Male Other & 4.30 \\
\hline
\hline
\end{tabular}
\caption{COMPAS dataset: statistics with reference to the sensitive features.}
\label{tab:COMPAS:statistics}
\end{table}

\begin{table}
\centering
\setlength{\tabcolsep}{0.06cm}
\renewcommand{\arraystretch}{1.7}
\begin{tabular}{|c||C{.6cm}|C{.6cm}|}
\hline
\hline
G & M & F \\ 
\hline
\hline
M & $58.2$ & $3.8$ \\
\hline
F & $8.7$ & $29.4$ \\
\hline
\hline
\end{tabular}\quad
\begin{tabular}{|c||C{.6cm}|C{.6cm}|C{.6cm}|C{.6cm}|C{.6cm}|}
\hline
\hline
R & W & B & API & AIE & O \\
\hline
\hline
W & $78.5$ & $1.7$ & $0.5$ & $0.2$ & $0.1$ \\
\hline
B & $4.6$ & $7.8$ & $0.1$ & $0.0$ & $0.0$ \\
\hline
API & $0.5$ & $0.0$ & $0.8$ & $0.0$ & $0.0$ \\
\hline
AIE & $1.5$ & $0.1$ & $0.0$ & $2.6$ & $0.0$ \\
\hline
O & $0.4$ & $0.0$ & $0.0$ & $0.0$ & $0.7$ \\
\hline
\hline
\end{tabular}
\caption{Adult Dataset: confusion matrices in percentage (true class in columns and predicted classes in rows) obtained by predicting Gender and Race from the other non-sensitive features using Random Forests.}
\label{tab:confusion}
\end{table}
\begin{table}
\centering
\setlength{\tabcolsep}{0.06cm}
\renewcommand{\arraystretch}{1.7}
\begin{tabular}{|c||C{.6cm}|C{.6cm}|}
\hline
\hline
G & M & F \\ 
\hline
\hline
M & $16.7$ & $8.6$ \\
\hline
F & $2.6$ & $72.1$ \\
\hline
\hline
\end{tabular}\quad
\begin{tabular}{|c||C{.6cm}|C{.6cm}|C{.6cm}|C{.6cm}|C{.6cm}|C{.6cm}|}
\hline
\hline
R & AA & A & C & H & NA & O \\
\hline
\hline
AA & $44.8$ & $0.0$ & $3.4$ & $0.6$ & $0.0$ & $0.3$ \\
\hline
A & $0.1$ & $0.3$ & $0.0$ & $0.0$ & $0.0$ & $0.0$ \\
\hline
C & $4.4$ & $0.0$ & $29.6$ & $0.4$ & $0.0$ & $0.2$ \\
\hline
H & $1.2$ & $0.0$ & $0.6$ & $7.7$ & $0.0$ & $0.1$ \\
\hline
NA & $0.0$ & $0.0$ & $0.0$ & $0.0$ & $0.2$ & $0.0$ \\
\hline
O & $0.7$ & $0.0$ & $0.4$ & $0.1$ & $0.0$ & $4.6$ \\
\hline
\hline
\end{tabular}
\caption{COMPAS Dataset: confusion matrices in percentage (true class in columns and predicted classes in rows) obtained by predicting Gender and Race from the other non-sensitive features using Random Forests.}
\label{tab:COMPASconfusion}
\end{table}

In all the experiments, we compare STL, ITL, and MTL in different settings. 
Specifically we test each method in the following cases: when the models use the sensitive feature (S${=}1$) or not (S${=}0$), when the fairness constraint is active (F${=}1$) or not (F${{=}}0$), when we consider the group specific models (D${=}1$) or the shared model between groups (D${=}0$), and when we use the true sensitive feature (P${=}1$) or the predicted one (P${=}0$).
Note that when D${=}0$ we can only compare STL with MTL, since only these two models produce a shared model between the groups, and furthermore, when D${=}1$ we can only compare ITL with MTL, since these produce group specific models. 

We collect statistics concerning the classification average accuracy per group in percentage (ACC) on the test set, difference of equal opportunities on both the positive and negative class (denoted as DEO$^+$ and DEO$^-$, respectively), and the difference of equalized odds (DEOd) of the selected model - see Section 2 for a definition of these quantities.

We selected the best hyperparameters\footnote{The ranges of hyperparameters used in the validation procedure of STL, MTL, and ITL are $\rho {\in} \{10^{-6.0}, 10^{-5.5}, \dots , 10^{+6.0} \}$ and $\lambda,\theta {\in} \{0, 2^{-15}, 2^{-14}, \dots , 2^{-1}, 1{-}2^{-2}, \dots , 1{-}2^{-15}, 1 \}$.} by 
the two steps 10-fold cross validation (CV) procedure described in~\cite{donini2018empirical}.
In the first step, the value of the hyperparameters with highest accuracy is identified.
In the second step, we shortlist all the hyperparameters with accuracy close to the best one (in our case, above $97\%$ of the best accuracy).
Finally, from this list, we select the hyperparameters with the lowest fairness measure.
This validation procedure, ensures that fairness cannot be achieved by a mere modification of hyperparameter selection procedure.

\subsection{Results}

\begin{figure}[!ht]
\centering
\includegraphics[width=.9\columnwidth]{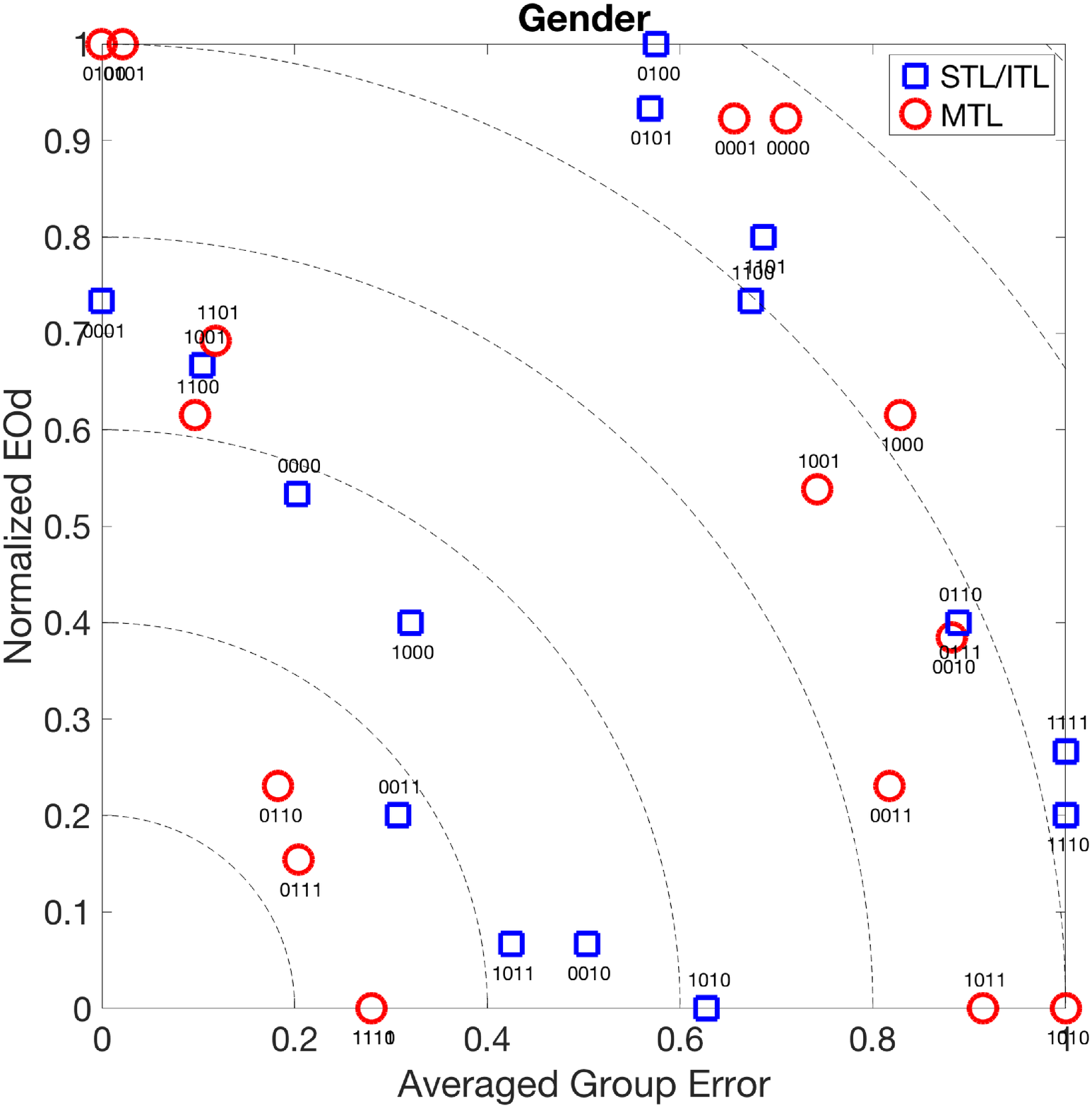}
\caption{Adult dataset: complete results set for Gender (text close to the symbols in plot are P, D, F, and S).}
\label{fig:ristot1}
\end{figure}

\begin{figure}[!ht]
\centering
\includegraphics[width=.9\columnwidth]{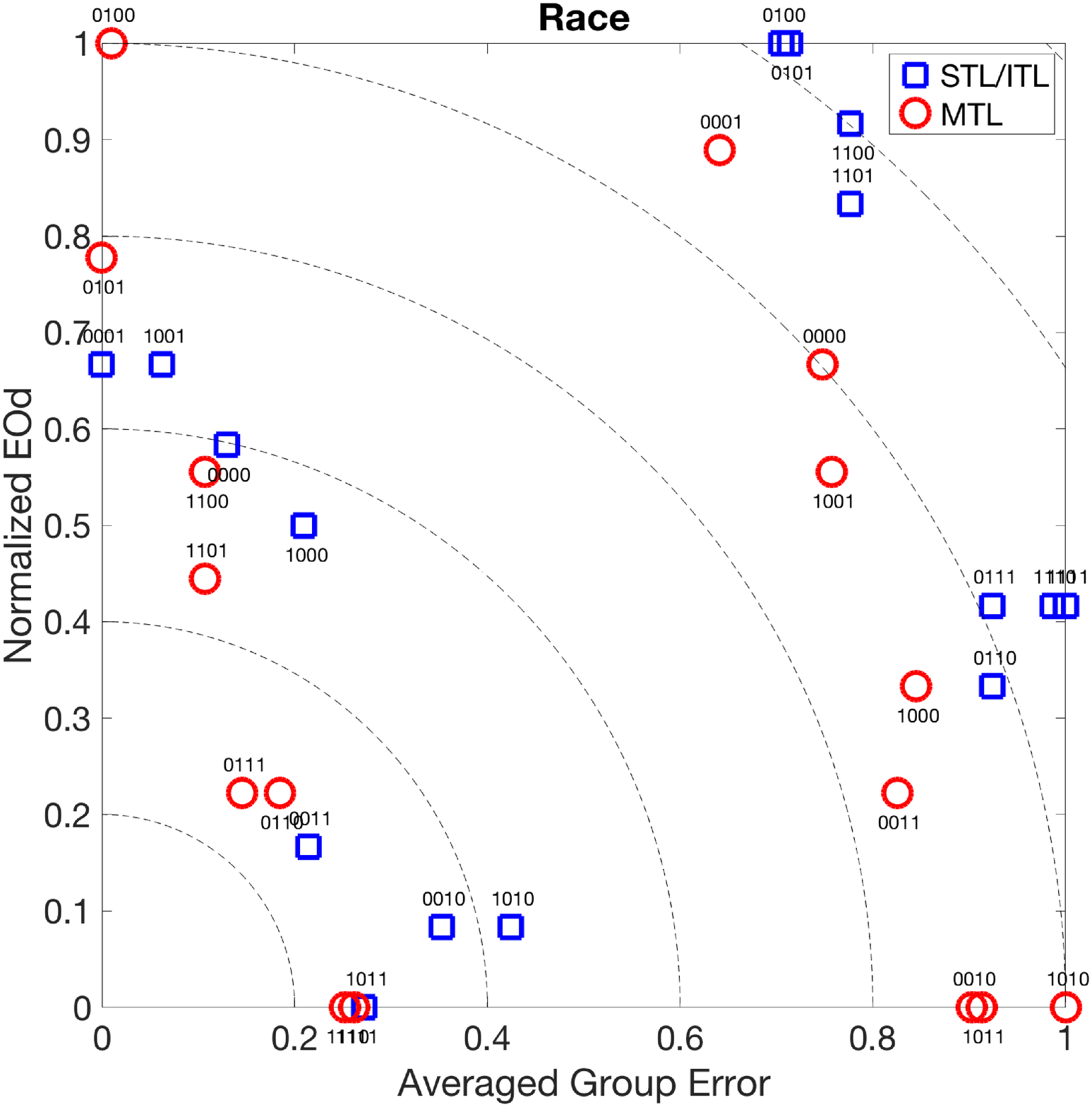} 
\caption{Adult dataset: complete results set for Race (text close to the symbols in plot are P, D, F, and S).}
\label{fig:ristot2}
\end{figure}

\begin{figure}[!ht]
\centering
\includegraphics[width=.9\columnwidth]{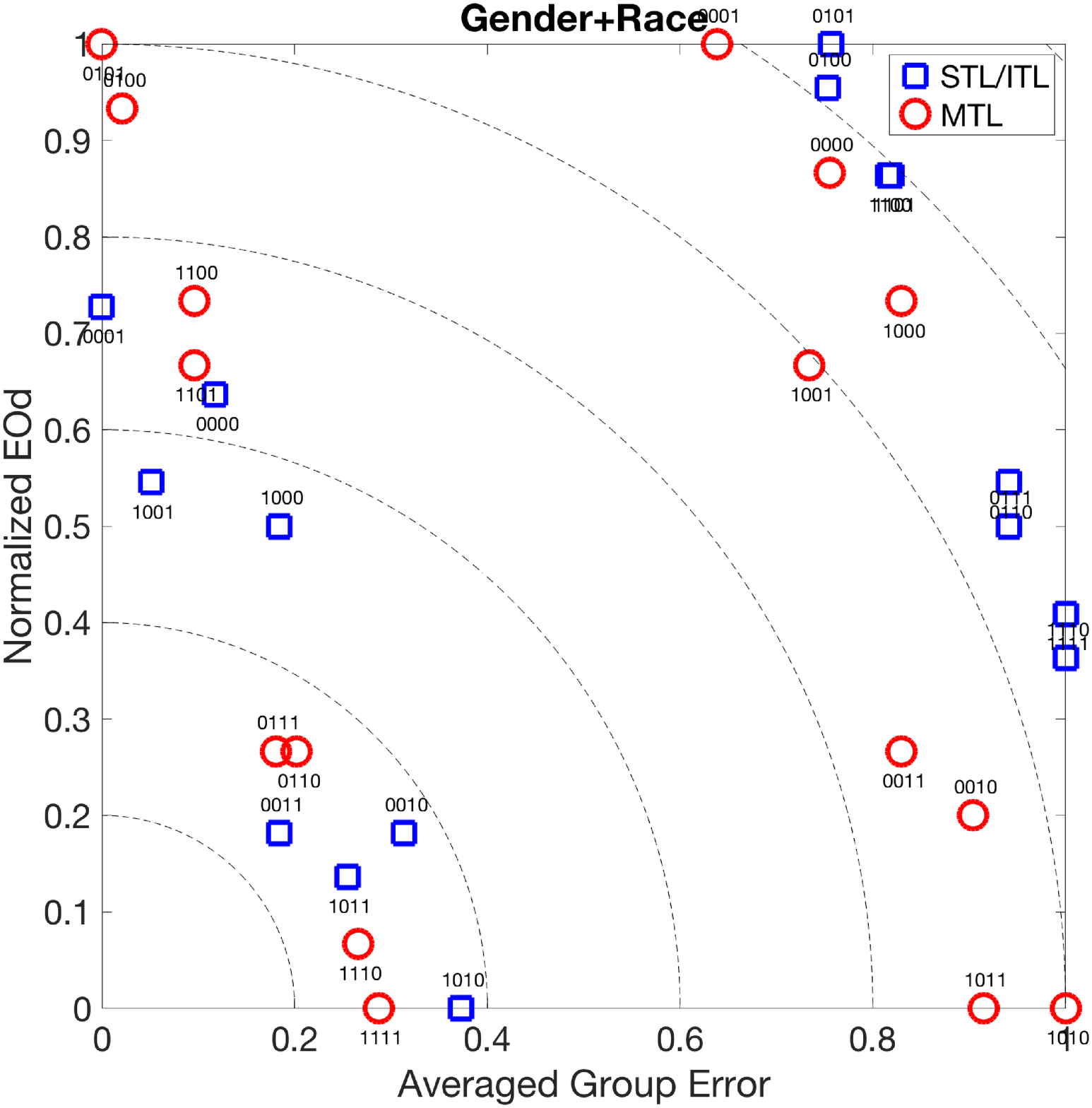}
\caption{Adult dataset: complete results set for Gender+Race (text close to the symbols in plot are P, D, F, and S).}
\label{fig:ristot3}
\end{figure}

The results for all possible combinations described above, are reported in Table~\ref{tab:ristot}.
In Figures~\ref{fig:ristot1},~\ref{fig:ristot2}, and~\ref{fig:ristot3}, we present a visualization of Table~\ref{tab:ristot} for the Adult dataset (results are analogous for the COMPAS one).
Where both the error (i.e., 1-ACC), and the EOd are normalized to be between 0 and 1, column-wise.
The closer a point is to the origin, the better the result.
\begin{table*}[h!]
\footnotesize
\centering
\setlength{\tabcolsep}{0.04cm}
\renewcommand{\arraystretch}{1.}
\begin{tabular}{||c||c|c|c|c||||c|c||c|c|||c|c||c|c|||c|c||c|c||||c|c||c|c|||c|c||c|c|||c|c||c|c||||}
\hline
\hline
\multicolumn{5}{||c||||}{} 
& \multicolumn{12}{c||||}{Adult Dataset}
& \multicolumn{12}{c||||}{COMPAS Dataset} \\
\cline{2-29}
& $-$ & $0$ & $-$ & $-$ 
& \multicolumn{2}{c||}{STL} & \multicolumn{2}{c|||}{\multirow{2}{*}{MTL}} 
& \multicolumn{2}{c||}{STL} & \multicolumn{2}{c|||}{\multirow{2}{*}{MTL}} 
& \multicolumn{2}{c||}{STL} & \multicolumn{2}{c||||}{\multirow{2}{*}{MTL}}
& \multicolumn{2}{c||}{STL} & \multicolumn{2}{c|||}{\multirow{2}{*}{MTL}} 
& \multicolumn{2}{c||}{STL} & \multicolumn{2}{c|||}{\multirow{2}{*}{MTL}} 
& \multicolumn{2}{c||}{STL} & \multicolumn{2}{c||||}{\multirow{2}{*}{MTL}}\\
& $-$ & $1$ & $-$ & $-$ 
& \multicolumn{2}{c||}{ITL} & \multicolumn{2}{c|||}{} 
& \multicolumn{2}{c||}{ITL} & \multicolumn{2}{c|||}{} 
& \multicolumn{2}{c||}{ITL} & \multicolumn{2}{c||||}{}
& \multicolumn{2}{c||}{ITL} & \multicolumn{2}{c|||}{} 
& \multicolumn{2}{c||}{ITL} & \multicolumn{2}{c|||}{} 
& \multicolumn{2}{c||}{ITL} & \multicolumn{2}{c||||}{}\\
\cline{2-29}
& P & D & F & S 
& \tiny ACC & \tiny DEOp$^+$ & \tiny ACC & \tiny DEOp$^+$ 
& \tiny ACC & \tiny DEOp$^-$ & \tiny ACC & \tiny DEOp$^-$ 
& \tiny \tiny ACC & \tiny DEOd & \tiny ACC & \tiny DEOd 
& \tiny ACC & \tiny DEOp$^+$ & \tiny ACC & \tiny DEOp$^+$ 
& \tiny ACC & \tiny DEOp$^-$ & \tiny ACC & \tiny DEOp$^-$ 
& \tiny \tiny ACC & \tiny DEOd & \tiny ACC & \tiny DEOd \\ 
\hline
\hline
\multirow{16}{*}{G}
& $0$ & $0$ & $0$ & $0$ & $80.2$ & $0.11$ & $83.4$ & $0.13$ & $80.4$ & $0.09$ & $84.3$ & $0.12$ & $80.3$ & $0.10$ & $83.6$ & $0.13$& $76.1$ & $0.15$ & $78.1$ & $0.12$ & $76.3$ & $0.14$ & $78.0$ & $0.11$ & $76.2$ & $0.13$ & $77.3$ & $0.10$\\ 
& $0$ & $0$ & $0$ & $1$ & $83.3$ & $0.14$ & $83.9$ & $0.13$ & $83.5$ & $0.12$ & $84.8$ & $0.12$ & $83.4$ & $0.13$ & $84.1$ & $0.13$& $79.3$ & $0.15$ & $79.2$ & $0.13$ & $79.5$ & $0.14$ & $79.1$ & $0.12$ & $79.4$ & $0.13$ & $78.4$ & $0.11$\\ 
& $0$ & $0$ & $1$ & $0$ & $75.7$ & $0.03$ & $81.8$ & $0.06$ & $75.8$ & $0.02$ & $82.7$ & $0.05$ & $75.7$ & $0.03$ & $82.0$ & $0.06$& $71.5$ & $0.03$ & $76.5$ & $0.03$ & $71.7$ & $0.03$ & $76.4$ & $0.03$ & $71.6$ & $0.03$ & $75.7$ & $0.03$\\ 
& $0$ & $0$ & $1$ & $1$ & $78.6$ & $0.06$ & $82.4$ & $0.04$ & $78.8$ & $0.05$ & $83.3$ & $0.04$ & $78.7$ & $0.05$ & $82.6$ & $0.04$& $74.4$ & $0.05$ & $77.4$ & $0.05$ & $74.6$ & $0.05$ & $77.3$ & $0.04$ & $74.5$ & $0.05$ & $76.6$ & $0.04$\\ 
& $0$ & $1$ & $0$ & $0$ & $74.5$ & $0.18$ & $90.0$ & $0.14$ & $74.7$ & $0.15$ & $91.0$ & $0.13$ & $74.6$ & $0.17$ & $90.2$ & $0.14$& $70.7$ & $0.19$ & $84.5$ & $0.15$ & $70.9$ & $0.17$ & $84.4$ & $0.14$ & $70.8$ & $0.16$ & $83.6$ & $0.13$\\ 
& $0$ & $1$ & $0$ & $1$ & $74.6$ & $0.17$ & $89.7$ & $0.14$ & $74.7$ & $0.15$ & $90.7$ & $0.13$ & $74.7$ & $0.16$ & $90.0$ & $0.14$& $70.9$ & $0.19$ & $84.5$ & $0.14$ & $71.1$ & $0.18$ & $84.4$ & $0.13$ & $71.0$ & $0.16$ & $83.6$ & $0.12$\\ 
& $0$ & $1$ & $1$ & $0$ & $69.7$ & $0.08$ & $88.3$ & $0.04$ & $69.9$ & $0.07$ & $89.2$ & $0.04$ & $69.8$ & $0.08$ & $88.5$ & $0.04$& $66.1$ & $0.08$ & $83.0$ & $0.04$ & $66.3$ & $0.08$ & $82.8$ & $0.04$ & $66.2$ & $0.07$ & $82.1$ & $0.04$\\ 
& $0$ & $1$ & $1$ & $1$ & $69.7$ & $0.08$ & $88.1$ & $0.03$ & $69.9$ & $0.07$ & $89.1$ & $0.03$ & $69.8$ & $0.08$ & $88.3$ & $0.03$& $66.1$ & $0.09$ & $82.9$ & $0.07$ & $66.3$ & $0.08$ & $82.8$ & $0.06$ & $66.2$ & $0.08$ & $82.1$ & $0.06$\\ 
& $1$ & $0$ & $0$ & $0$ & $78.4$ & $0.09$ & $82.3$ & $0.09$ & $78.6$ & $0.07$ & $83.2$ & $0.09$ & $78.5$ & $0.08$ & $82.5$ & $0.09$& $74.6$ & $0.12$ & $77.3$ & $0.10$ & $74.8$ & $0.11$ & $77.2$ & $0.09$ & $74.7$ & $0.10$ & $76.5$ & $0.09$\\ 
& $1$ & $0$ & $0$ & $1$ & $81.7$ & $0.13$ & $83.1$ & $0.08$ & $81.9$ & $0.11$ & $84.0$ & $0.07$ & $81.8$ & $0.12$ & $83.3$ & $0.08$& $77.6$ & $0.13$ & $78.1$ & $0.09$ & $77.8$ & $0.12$ & $78.0$ & $0.09$ & $77.7$ & $0.11$ & $77.3$ & $0.08$\\ 
& $1$ & $0$ & $1$ & $0$ & $73.7$ & $0.02$ & $80.7$ & $0.01$ & $73.9$ & $0.02$ & $81.6$ & $0.01$ & $73.8$ & $0.02$ & $80.9$ & $0.01$& $70.1$ & $0.03$ & $75.9$ & $0.01$ & $70.3$ & $0.03$ & $75.8$ & $0.01$ & $70.2$ & $0.03$ & $75.1$ & $0.01$\\ 
& $1$ & $0$ & $1$ & $1$ & $76.8$ & $0.03$ & $81.5$ & $0.01$ & $77.0$ & $0.03$ & $82.4$ & $0.01$ & $76.9$ & $0.03$ & $81.7$ & $0.01$& $73.1$ & $0.05$ & $76.7$ & $0.01$ & $73.3$ & $0.05$ & $76.6$ & $0.01$ & $73.2$ & $0.04$ & $75.9$ & $0.01$\\ 
& $1$ & $1$ & $0$ & $0$ & $73.0$ & $0.14$ & $89.1$ & $0.09$ & $73.2$ & $0.12$ & $90.1$ & $0.08$ & $73.1$ & $0.13$ & $89.3$ & $0.09$& $69.3$ & $0.17$ & $83.7$ & $0.09$ & $69.5$ & $0.15$ & $83.6$ & $0.08$ & $69.4$ & $0.14$ & $82.8$ & $0.08$\\ 
& $1$ & $1$ & $0$ & $1$ & $72.8$ & $0.15$ & $88.9$ & $0.10$ & $73.0$ & $0.13$ & $89.9$ & $0.09$ & $72.9$ & $0.14$ & $89.1$ & $0.10$& $69.3$ & $0.15$ & $83.7$ & $0.10$ & $69.5$ & $0.14$ & $83.6$ & $0.09$ & $69.4$ & $0.13$ & $82.9$ & $0.09$\\ 
& $1$ & $1$ & $1$ & $0$ & $68.0$ & $0.06$ & $87.4$ & $0.01$ & $68.2$ & $0.05$ & $88.3$ & $0.01$ & $68.1$ & $0.05$ & $87.6$ & $0.01$& $64.7$ & $0.06$ & $82.3$ & $0.01$ & $64.9$ & $0.05$ & $82.1$ & $0.01$ & $64.8$ & $0.05$ & $81.4$ & $0.01$\\ 
& $1$ & $1$ & $1$ & $1$ & $68.0$ & $0.06$ & $87.4$ & $0.01$ & $68.1$ & $0.05$ & $88.3$ & $0.01$ & $68.1$ & $0.06$ & $87.6$ & $0.01$& $64.6$ & $0.06$ & $82.1$ & $0.01$ & $64.8$ & $0.06$ & $82.0$ & $0.01$ & $64.7$ & $0.05$ & $81.3$ & $0.01$\\ 
\hline
\hline
\multirow{16}{*}{R}
& $0$ & $0$ & $0$ & $0$ & $80.3$ & $0.08$ & $84.2$ & $0.07$ & $80.5$ & $0.07$ & $85.1$ & $0.06$ & $80.4$ & $0.08$ & $84.4$ & $0.07$& $80.2$ & $0.09$ & $84.2$ & $0.08$ & $80.4$ & $0.08$ & $85.1$ & $0.07$ & $80.3$ & $0.09$ & $84.4$ & $0.08$\\ 
& $0$ & $0$ & $0$ & $1$ & $83.2$ & $0.09$ & $85.3$ & $0.09$ & $83.4$ & $0.08$ & $86.2$ & $0.08$ & $83.3$ & $0.09$ & $85.5$ & $0.09$& $83.2$ & $0.10$ & $84.9$ & $0.08$ & $83.4$ & $0.09$ & $85.8$ & $0.07$ & $83.3$ & $0.10$ & $85.1$ & $0.08$\\ 
& $0$ & $0$ & $1$ & $0$ & $75.3$ & $0.02$ & $82.6$ & $0.01$ & $75.5$ & $0.02$ & $83.5$ & $0.01$ & $75.4$ & $0.02$ & $82.8$ & $0.01$& $75.5$ & $0.04$ & $82.4$ & $0.03$ & $75.7$ & $0.04$ & $83.3$ & $0.03$ & $75.6$ & $0.04$ & $82.6$ & $0.03$\\ 
& $0$ & $0$ & $1$ & $1$ & $78.4$ & $0.03$ & $83.4$ & $0.03$ & $78.6$ & $0.03$ & $84.3$ & $0.02$ & $78.5$ & $0.03$ & $83.6$ & $0.03$& $78.5$ & $0.05$ & $83.5$ & $0.02$ & $78.7$ & $0.04$ & $84.4$ & $0.02$ & $78.6$ & $0.05$ & $83.7$ & $0.02$\\ 
& $0$ & $1$ & $0$ & $0$ & $67.4$ & $0.13$ & $91.8$ & $0.10$ & $67.6$ & $0.11$ & $92.8$ & $0.08$ & $67.5$ & $0.13$ & $92.0$ & $0.10$& $67.3$ & $0.12$ & $91.7$ & $0.08$ & $67.5$ & $0.11$ & $92.7$ & $0.07$ & $67.4$ & $0.12$ & $92.0$ & $0.08$\\ 
& $0$ & $1$ & $0$ & $1$ & $67.2$ & $0.13$ & $91.8$ & $0.08$ & $67.4$ & $0.12$ & $92.8$ & $0.07$ & $67.3$ & $0.13$ & $92.1$ & $0.08$& $67.4$ & $0.13$ & $91.8$ & $0.09$ & $67.5$ & $0.11$ & $92.8$ & $0.08$ & $67.4$ & $0.13$ & $92.0$ & $0.09$\\ 
& $0$ & $1$ & $1$ & $0$ & $62.5$ & $0.05$ & $90.0$ & $0.03$ & $62.7$ & $0.05$ & $90.9$ & $0.03$ & $62.6$ & $0.05$ & $90.2$ & $0.03$& $62.4$ & $0.07$ & $90.1$ & $0.02$ & $62.6$ & $0.06$ & $91.0$ & $0.02$ & $62.5$ & $0.07$ & $90.3$ & $0.02$\\ 
& $0$ & $1$ & $1$ & $1$ & $62.6$ & $0.06$ & $90.4$ & $0.03$ & $62.7$ & $0.05$ & $91.3$ & $0.03$ & $62.6$ & $0.06$ & $90.6$ & $0.03$& $62.4$ & $0.07$ & $90.0$ & $0.03$ & $62.5$ & $0.07$ & $91.0$ & $0.03$ & $62.4$ & $0.07$ & $90.2$ & $0.03$\\ 
& $1$ & $0$ & $0$ & $0$ & $78.5$ & $0.07$ & $83.2$ & $0.04$ & $78.7$ & $0.06$ & $84.1$ & $0.04$ & $78.6$ & $0.07$ & $83.4$ & $0.04$& $78.4$ & $0.08$ & $83.3$ & $0.06$ & $78.6$ & $0.07$ & $84.2$ & $0.05$ & $78.5$ & $0.08$ & $83.5$ & $0.06$\\ 
& $1$ & $0$ & $0$ & $1$ & $81.8$ & $0.09$ & $84.1$ & $0.06$ & $82.0$ & $0.08$ & $85.0$ & $0.05$ & $81.9$ & $0.09$ & $84.3$ & $0.06$& $81.7$ & $0.09$ & $84.4$ & $0.07$ & $81.9$ & $0.08$ & $85.3$ & $0.06$ & $81.8$ & $0.09$ & $84.6$ & $0.07$\\ 
& $1$ & $0$ & $1$ & $0$ & $73.7$ & $0.02$ & $81.6$ & $0.01$ & $73.9$ & $0.02$ & $82.5$ & $0.01$ & $73.8$ & $0.02$ & $81.8$ & $0.01$& $73.7$ & $0.01$ & $81.5$ & $0.01$ & $73.9$ & $0.01$ & $82.4$ & $0.01$ & $73.8$ & $0.01$ & $81.7$ & $0.01$\\ 
& $1$ & $0$ & $1$ & $1$ & $77.1$ & $0.01$ & $82.5$ & $0.01$ & $77.3$ & $0.01$ & $83.4$ & $0.01$ & $77.2$ & $0.01$ & $82.7$ & $0.01$& $77.0$ & $0.02$ & $82.4$ & $0.01$ & $77.2$ & $0.01$ & $83.2$ & $0.01$ & $77.1$ & $0.02$ & $82.5$ & $0.01$\\ 
& $1$ & $1$ & $0$ & $0$ & $65.8$ & $0.12$ & $90.8$ & $0.06$ & $66.0$ & $0.11$ & $91.8$ & $0.05$ & $65.9$ & $0.12$ & $91.0$ & $0.06$& $65.5$ & $0.12$ & $90.8$ & $0.05$ & $65.7$ & $0.11$ & $91.8$ & $0.05$ & $65.6$ & $0.12$ & $91.0$ & $0.05$\\ 
& $1$ & $1$ & $0$ & $1$ & $65.8$ & $0.11$ & $90.7$ & $0.05$ & $66.0$ & $0.10$ & $91.7$ & $0.04$ & $65.9$ & $0.11$ & $91.0$ & $0.05$& $65.7$ & $0.12$ & $90.8$ & $0.07$ & $65.8$ & $0.11$ & $91.7$ & $0.07$ & $65.7$ & $0.12$ & $91.0$ & $0.07$\\
& $1$ & $1$ & $1$ & $0$ & $61.2$ & $0.06$ & $89.3$ & $0.01$ & $61.3$ & $0.05$ & $90.3$ & $0.01$ & $61.2$ & $0.06$ & $89.5$ & $0.01$& $60.8$ & $0.05$ & $89.2$ & $0.01$ & $61.0$ & $0.05$ & $90.1$ & $0.01$ & $60.9$ & $0.05$ & $89.4$ & $0.01$\\ 
& $1$ & $1$ & $1$ & $1$ & $60.8$ & $0.06$ & $89.2$ & $0.01$ & $61.0$ & $0.05$ & $90.2$ & $0.01$ & $60.9$ & $0.06$ & $89.4$ & $0.01$& $60.9$ & $0.04$ & $89.0$ & $0.01$ & $61.1$ & $0.04$ & $89.9$ & $0.01$ & $61.0$ & $0.04$ & $89.2$ & $0.01$\\ 
\hline
\hline
\multirow{16}{*}{\tiny G+R}
& $0$ & $0$ & $0$ & $0$ & $80.2$ & $0.16$ & $84.6$ & $0.14$ & $80.4$ & $0.14$ & $85.3$ & $0.14$ & $80.3$ & $0.15$ & $84.9$ & $0.14$& $80.2$ & $0.16$ & $84.8$ & $0.14$ & $80.4$ & $0.14$ & $85.5$ & $0.14$ & $80.3$ & $0.15$ & $85.1$ & $0.14$\\ 
& $0$ & $0$ & $0$ & $1$ & $83.1$ & $0.18$ & $85.7$ & $0.16$ & $83.4$ & $0.16$ & $86.4$ & $0.16$ & $83.3$ & $0.17$ & $86.0$ & $0.16$& $83.3$ & $0.18$ & $85.5$ & $0.16$ & $83.5$ & $0.15$ & $86.2$ & $0.16$ & $83.4$ & $0.16$ & $85.8$ & $0.16$\\ 
& $0$ & $0$ & $1$ & $0$ & $75.2$ & $0.05$ & $83.2$ & $0.04$ & $75.3$ & $0.04$ & $83.9$ & $0.04$ & $75.3$ & $0.05$ & $83.5$ & $0.04$& $75.3$ & $0.05$ & $83.1$ & $0.05$ & $75.5$ & $0.04$ & $83.8$ & $0.05$ & $75.4$ & $0.05$ & $83.4$ & $0.05$\\ 
& $0$ & $0$ & $1$ & $1$ & $78.5$ & $0.05$ & $83.9$ & $0.05$ & $78.7$ & $0.04$ & $84.6$ & $0.05$ & $78.6$ & $0.05$ & $84.2$ & $0.05$& $78.6$ & $0.06$ & $84.1$ & $0.04$ & $78.8$ & $0.05$ & $84.7$ & $0.04$ & $78.7$ & $0.06$ & $84.3$ & $0.04$\\ 
& $0$ & $1$ & $0$ & $0$ & $64.0$ & $0.23$ & $91.5$ & $0.15$ & $64.2$ & $0.20$ & $92.2$ & $0.15$ & $64.1$ & $0.22$ & $91.8$ & $0.15$& $64.2$ & $0.24$ & $91.4$ & $0.16$ & $64.3$ & $0.21$ & $92.2$ & $0.16$ & $64.3$ & $0.22$ & $91.7$ & $0.16$\\ 
& $0$ & $1$ & $0$ & $1$ & $63.9$ & $0.24$ & $91.7$ & $0.16$ & $64.0$ & $0.21$ & $92.4$ & $0.16$ & $64.0$ & $0.23$ & $92.0$ & $0.16$& $64.1$ & $0.23$ & $91.5$ & $0.15$ & $64.3$ & $0.20$ & $92.2$ & $0.15$ & $64.2$ & $0.22$ & $91.8$ & $0.15$\\ 
& $0$ & $1$ & $1$ & $0$ & $59.3$ & $0.14$ & $89.8$ & $0.05$ & $59.4$ & $0.12$ & $90.6$ & $0.05$ & $59.3$ & $0.13$ & $90.1$ & $0.05$& $59.2$ & $0.13$ & $90.1$ & $0.05$ & $59.4$ & $0.11$ & $90.8$ & $0.05$ & $59.3$ & $0.12$ & $90.4$ & $0.05$\\ 
& $0$ & $1$ & $1$ & $1$ & $59.2$ & $0.13$ & $90.0$ & $0.05$ & $59.4$ & $0.11$ & $90.8$ & $0.05$ & $59.3$ & $0.12$ & $90.3$ & $0.05$& $59.4$ & $0.13$ & $89.9$ & $0.05$ & $59.5$ & $0.11$ & $90.7$ & $0.05$ & $59.5$ & $0.12$ & $90.3$ & $0.05$\\ 
& $1$ & $0$ & $0$ & $0$ & $78.5$ & $0.13$ & $83.9$ & $0.12$ & $78.7$ & $0.11$ & $84.6$ & $0.12$ & $78.6$ & $0.12$ & $84.2$ & $0.12$& $78.4$ & $0.13$ & $83.8$ & $0.09$ & $78.6$ & $0.12$ & $84.5$ & $0.09$ & $78.5$ & $0.13$ & $84.1$ & $0.09$\\ 
& $1$ & $0$ & $0$ & $1$ & $81.9$ & $0.14$ & $84.8$ & $0.11$ & $82.1$ & $0.12$ & $85.5$ & $0.11$ & $82.0$ & $0.13$ & $85.1$ & $0.11$& $81.7$ & $0.15$ & $84.7$ & $0.11$ & $81.9$ & $0.13$ & $85.4$ & $0.11$ & $81.8$ & $0.14$ & $85.0$ & $0.11$\\
& $1$ & $0$ & $1$ & $0$ & $73.7$ & $0.01$ & $82.3$ & $0.01$ & $73.9$ & $0.01$ & $83.0$ & $0.01$ & $73.8$ & $0.01$ & $82.6$ & $0.01$& $73.6$ & $0.02$ & $82.5$ & $0.01$ & $73.8$ & $0.02$ & $83.1$ & $0.01$ & $73.7$ & $0.02$ & $82.8$ & $0.01$\\ 
& $1$ & $0$ & $1$ & $1$ & $76.8$ & $0.04$ & $83.1$ & $0.01$ & $76.9$ & $0.04$ & $83.8$ & $0.01$ & $76.8$ & $0.04$ & $83.4$ & $0.01$& $76.8$ & $0.04$ & $83.2$ & $0.01$ & $77.0$ & $0.04$ & $83.8$ & $0.01$ & $76.9$ & $0.04$ & $83.4$ & $0.01$\\ 
& $1$ & $1$ & $0$ & $0$ & $62.5$ & $0.21$ & $90.8$ & $0.12$ & $62.6$ & $0.19$ & $91.5$ & $0.12$ & $62.5$ & $0.20$ & $91.1$ & $0.12$& $62.5$ & $0.21$ & $90.6$ & $0.11$ & $62.6$ & $0.18$ & $91.4$ & $0.11$ & $62.6$ & $0.20$ & $91.0$ & $0.11$\\ 
& $1$ & $1$ & $0$ & $1$ & $62.3$ & $0.21$ & $90.8$ & $0.11$ & $62.5$ & $0.18$ & $91.5$ & $0.11$ & $62.4$ & $0.20$ & $91.1$ & $0.11$ & $62.5$ & $0.22$ & $90.7$ & $0.11$ & $62.6$ & $0.19$ & $91.4$ & $0.11$ & $62.6$ & $0.20$ & $91.0$ & $0.11$\\ 
& $1$ & $1$ & $1$ & $0$ & $57.7$ & $0.10$ & $89.1$ & $0.02$ & $57.9$ & $0.08$ & $89.9$ & $0.02$ & $57.8$ & $0.09$ & $89.5$ & $0.02$& $57.8$ & $0.11$ & $89.2$ & $0.01$ & $58.0$ & $0.10$ & $89.9$ & $0.01$ & $57.9$ & $0.11$ & $89.5$ & $0.01$\\ 
& $1$ & $1$ & $1$ & $1$ & $57.7$ & $0.10$ & $89.0$ & $0.01$ & $57.9$ & $0.09$ & $89.8$ & $0.01$ & $57.8$ & $0.10$ & $89.3$ & $0.01$& $57.7$ & $0.10$ & $89.0$ & $0.01$ & $57.8$ & $0.08$ & $89.8$ & $0.01$ & $57.7$ & $0.09$ & $89.3$ & $0.01$\\ 
\hline
\hline
\end{tabular}
\caption{Complete results set.}
\label{tab:ristot}
\end{table*}

\begin{figure*}
%\vspace{-.25cm}
\centering
\includegraphics[trim={0.0cm 2.0cm 0.0cm 0.0cm}, width=0.65\columnwidth]{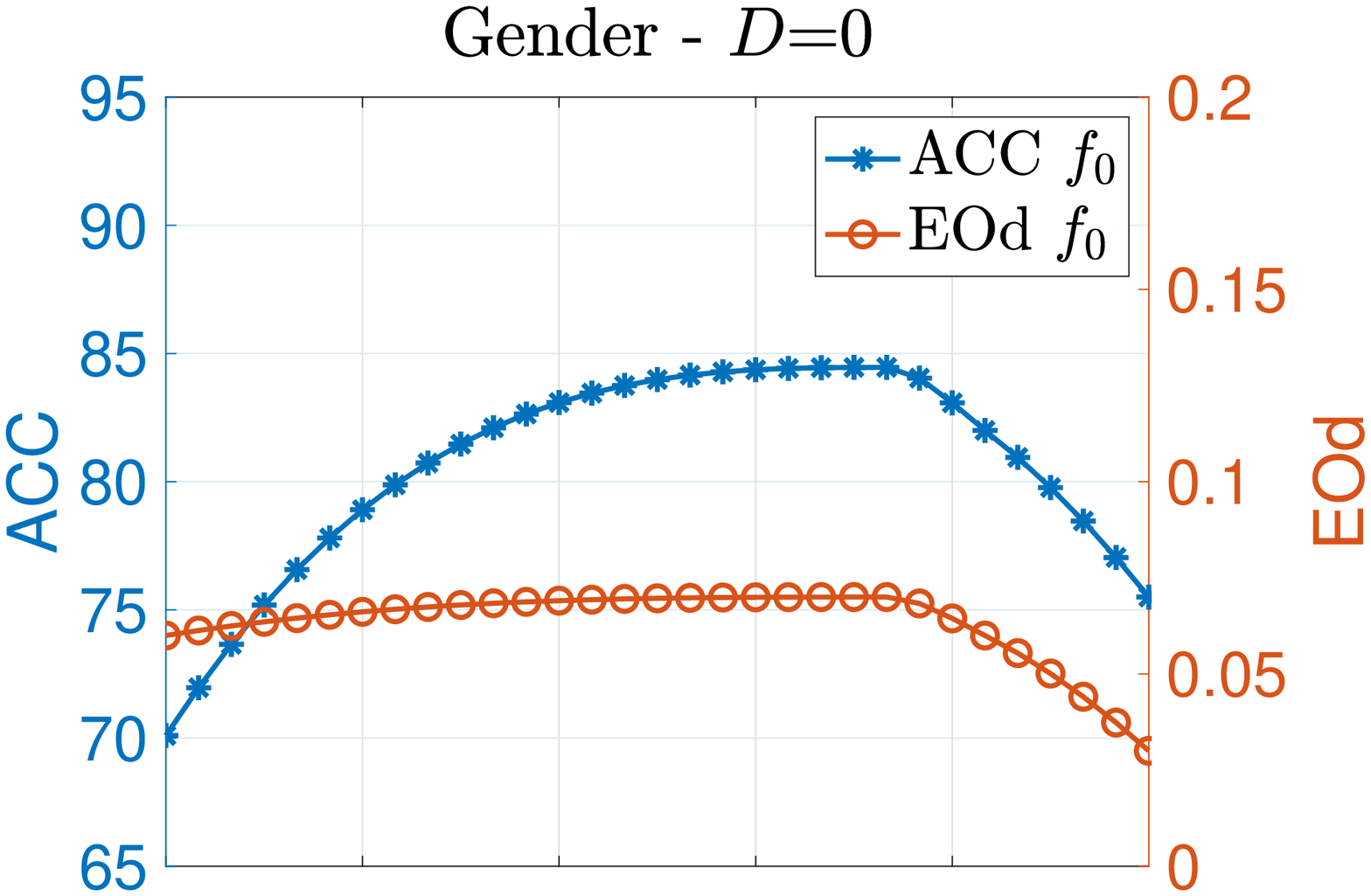} \quad
\includegraphics[trim={0.0cm 2.0cm 0.0cm 0.0cm}, width=0.65\columnwidth]{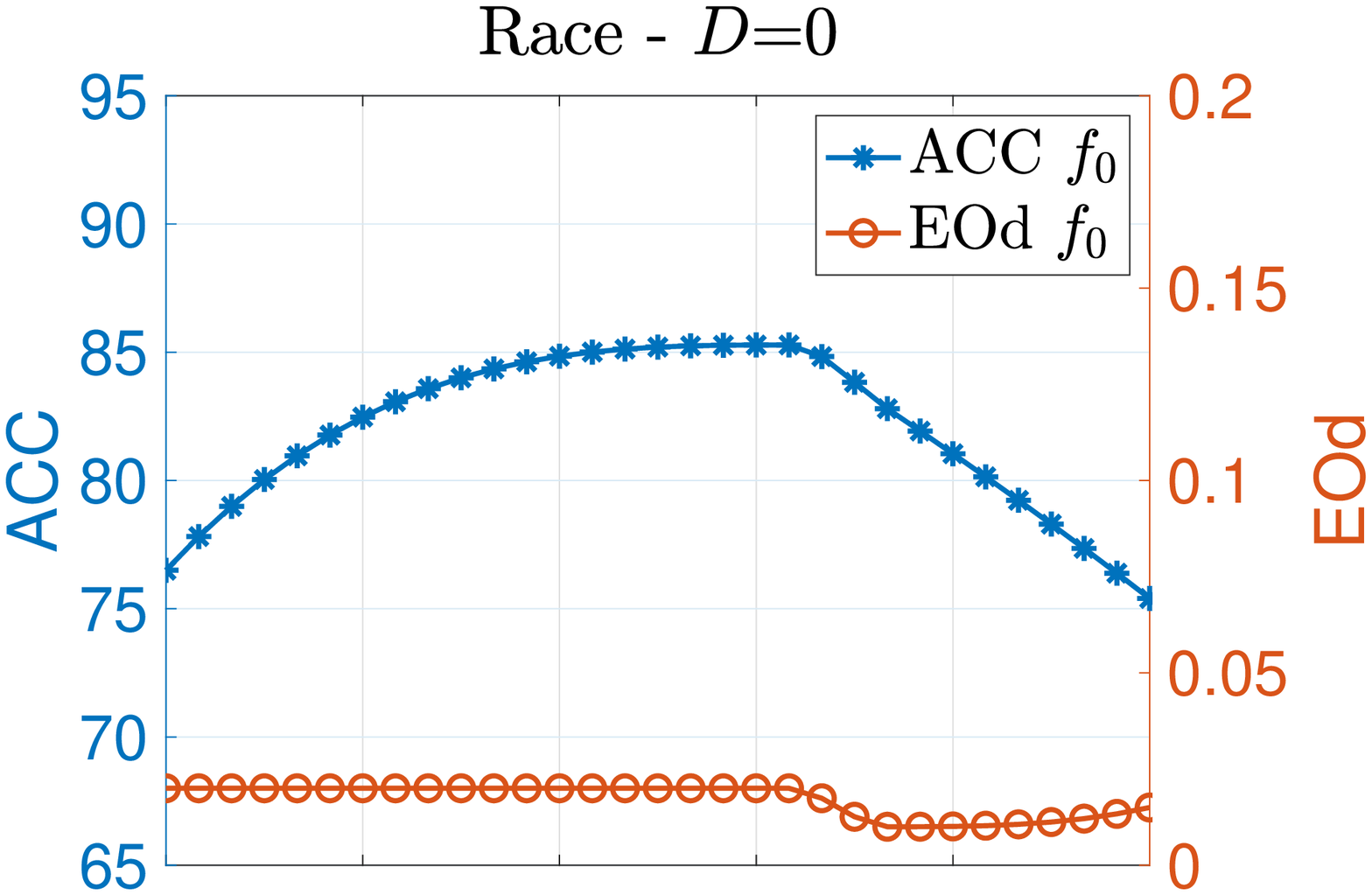} \quad
\includegraphics[trim={0.0cm 2.0cm 0.0cm 0.0cm}, width=0.65\columnwidth]{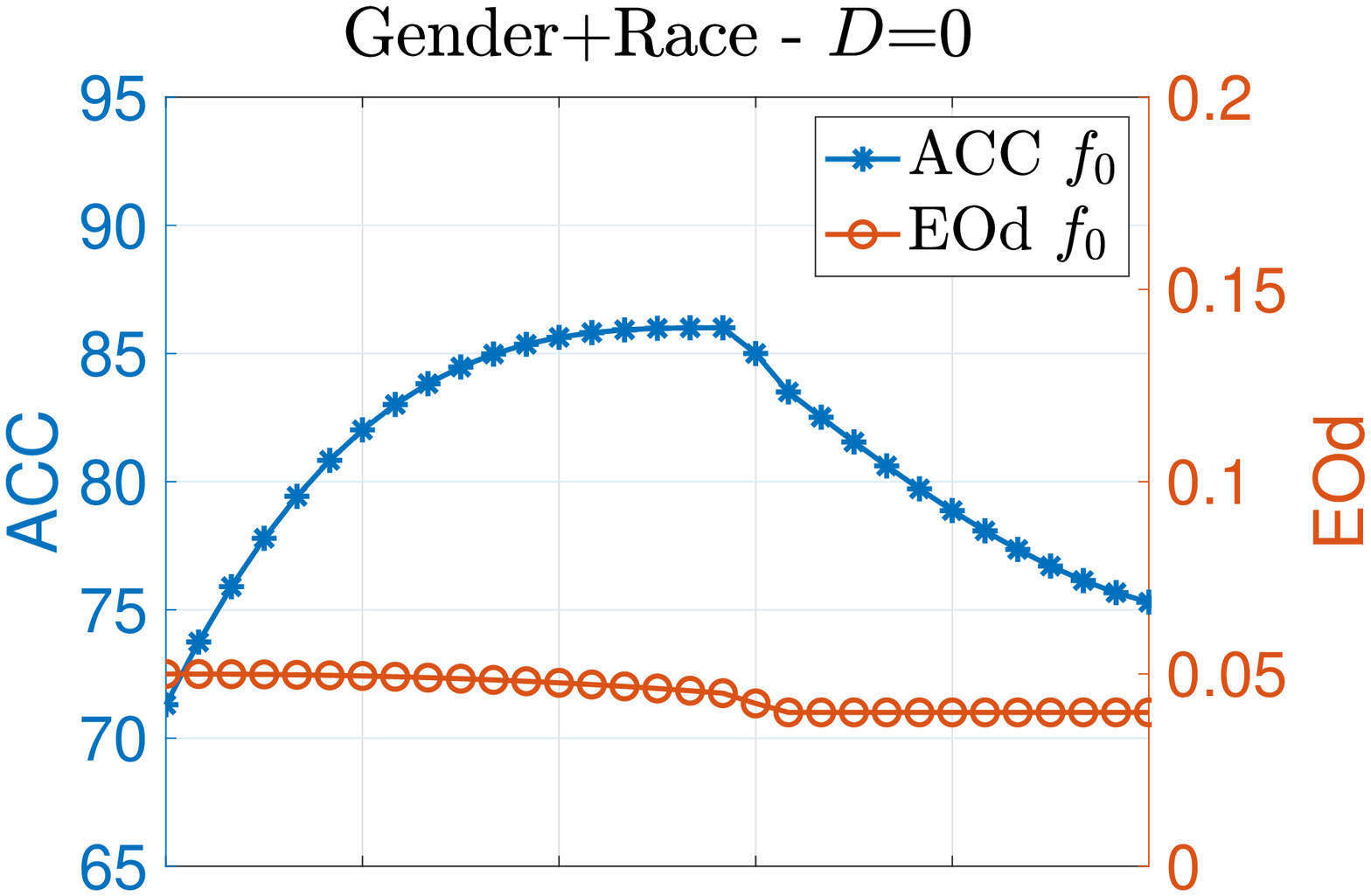} \\
\includegraphics[trim={0.0cm 0.0cm 0.0cm 0.0cm}, width=0.65\columnwidth]{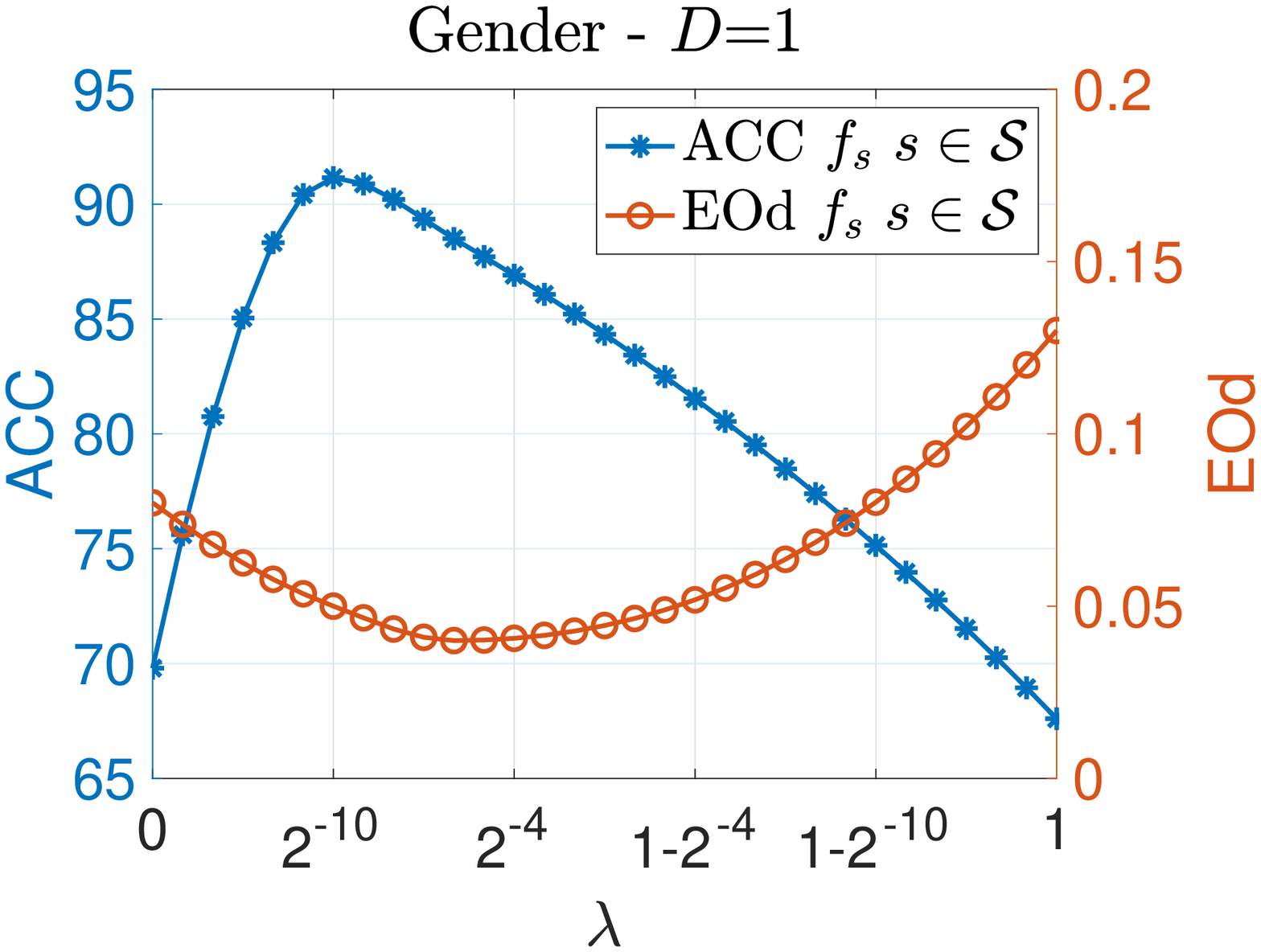} \quad
\includegraphics[trim={0.0cm 0.0cm 0.0cm 0.0cm}, width=0.65\columnwidth]{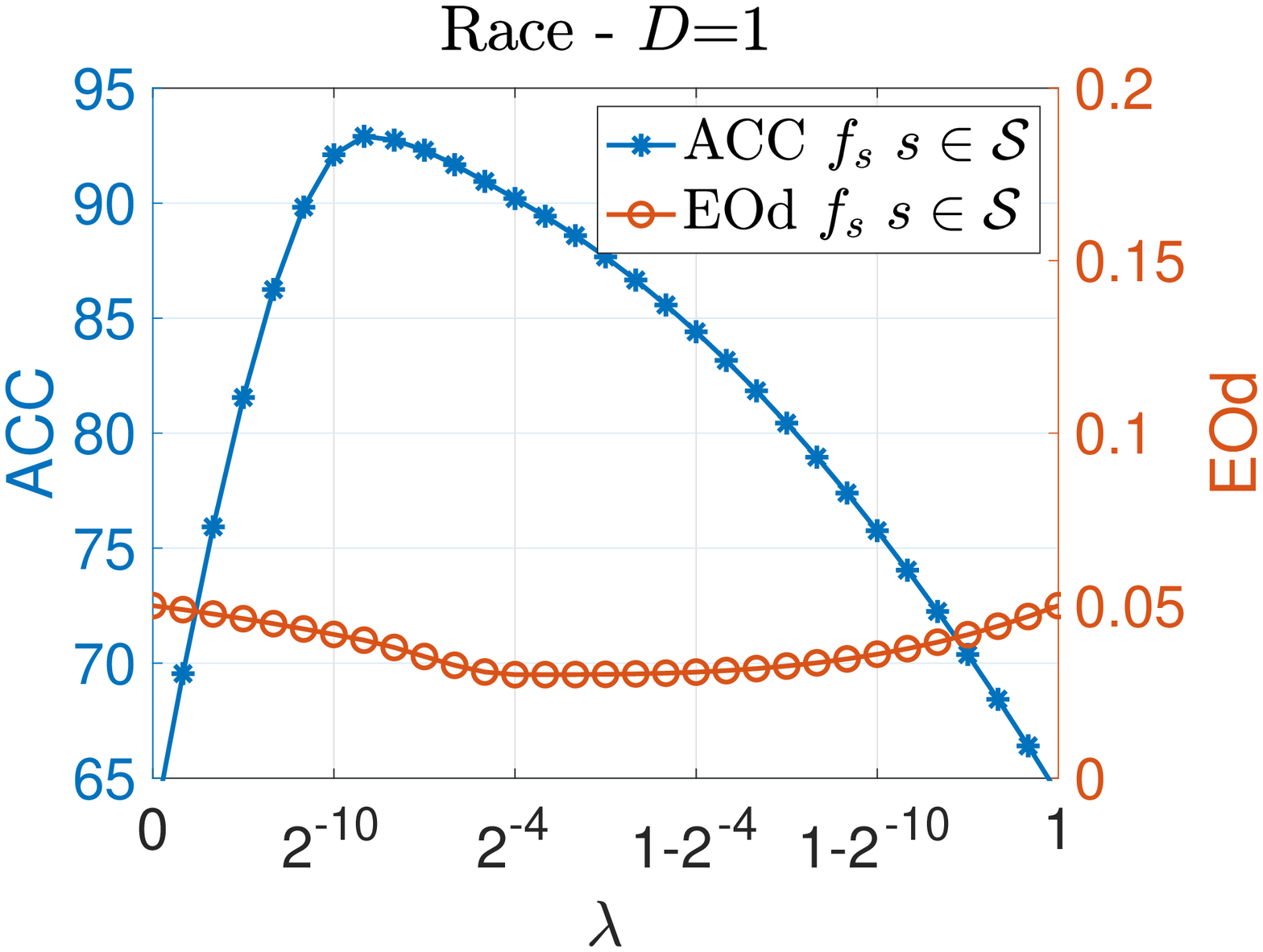} \quad
\includegraphics[trim={0.0cm 0.0cm 0.0cm 0.0cm}, width=0.65\columnwidth]{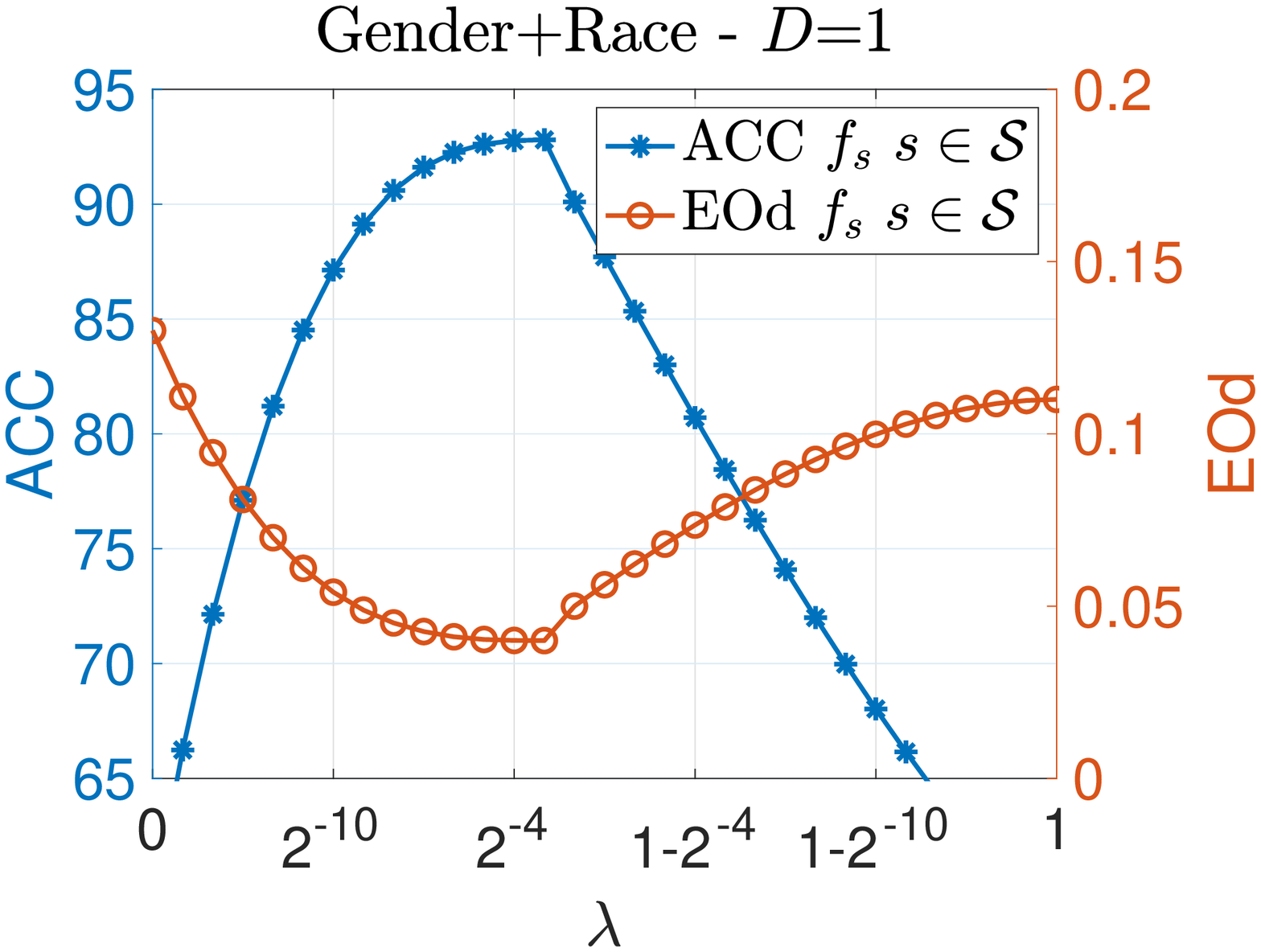} \\
\caption{Adult Dataset: ACC and EOd of MTL, when we fix
$\theta$ and $\rho$ to be the best values found during the validation procedure 
%$\theta{=}\theta^*$ and $\rho{=}\rho^*$ 
and we vary $\lambda$ with $P{=}0$, $F{=}1$, and $S{=}0$.
%\AM{Adult Dataset: ACC and EOd of MTL, for which we fix $\theta{=}\theta^*$, $\rho{=}\rho^*$. We vary $\lambda$, and set $P{=}0$, $F{=}1$, and $S{=}0$}
}
\label{fig:tableexplanation}
%\vspace{-.25cm}
\end{figure*}

Table~\ref{tab:ris1} presents the performance of the shared model trained with STL or MTL, with or without the sensitive feature as a predictor, and with or without the fairness constraint.
\begin{table*}[!h]
\footnotesize
\centering
\setlength{\tabcolsep}{0.04cm}
\renewcommand{\arraystretch}{1.}
\begin{tabular}{||c||c|c|c|c||||c|c||c|c|||c|c||c|c|||c|c||c|c||||c|c||c|c|||c|c||c|c|||c|c||c|c||||}
\hline
\hline
\multicolumn{5}{||c||||}{} 
& \multicolumn{12}{c||||}{Adult Dataset}
& \multicolumn{12}{c||||}{COMPAS Dataset} \\
\cline{2-29}
& $-$ & $0$ & $-$ & $-$ 
& \multicolumn{2}{c||}{STL} & \multicolumn{2}{c|||}{\multirow{2}{*}{MTL}} 
& \multicolumn{2}{c||}{STL} & \multicolumn{2}{c|||}{\multirow{2}{*}{MTL}} 
& \multicolumn{2}{c||}{STL} & \multicolumn{2}{c||||}{\multirow{2}{*}{MTL}}
& \multicolumn{2}{c||}{STL} & \multicolumn{2}{c|||}{\multirow{2}{*}{MTL}} 
& \multicolumn{2}{c||}{STL} & \multicolumn{2}{c|||}{\multirow{2}{*}{MTL}} 
& \multicolumn{2}{c||}{STL} & \multicolumn{2}{c||||}{\multirow{2}{*}{MTL}}\\
& $-$ & $1$ & $-$ & $-$ 
& \multicolumn{2}{c||}{ITL} & \multicolumn{2}{c|||}{} 
& \multicolumn{2}{c||}{ITL} & \multicolumn{2}{c|||}{} 
& \multicolumn{2}{c||}{ITL} & \multicolumn{2}{c||||}{}
& \multicolumn{2}{c||}{ITL} & \multicolumn{2}{c|||}{} 
& \multicolumn{2}{c||}{ITL} & \multicolumn{2}{c|||}{} 
& \multicolumn{2}{c||}{ITL} & \multicolumn{2}{c||||}{}\\
\cline{2-29}
& P & D & F & S 
& \tiny ACC & \tiny DEOp$^+$ & \tiny ACC & \tiny DEOp$^+$ 
& \tiny ACC & \tiny DEOp$^-$ & \tiny ACC & \tiny DEOp$^-$ 
& \tiny \tiny ACC & \tiny DEOd & \tiny ACC & \tiny DEOd 
& \tiny ACC & \tiny DEOp$^+$ & \tiny ACC & \tiny DEOp$^+$ 
& \tiny ACC & \tiny DEOp$^-$ & \tiny ACC & \tiny DEOp$^-$ 
& \tiny \tiny ACC & \tiny DEOd & \tiny ACC & \tiny DEOd \\ 
\hline
\hline
\multirow{3}{*}{G}
& $0$ & $0$ & $0$ & $0$ & $80.2$ & $0.11$ & $83.4$ & $0.13$ & $80.4$ & $0.09$ & $84.3$ & $0.12$ & $80.3$ & $0.10$ & $83.6$ & $0.13$& $76.1$ & $0.15$ & $78.1$ & $0.12$ & $76.3$ & $0.14$ & $78.0$ & $0.11$ & $76.2$ & $0.13$ & $77.3$ & $0.10$\\ 
& $0$ & $0$ & $1$ & $0$ & $75.7$ & $0.03$ & $81.8$ & $0.06$ & $75.8$ & $0.02$ & $82.7$ & $0.05$ & $75.7$ & $0.03$ & $82.0$ & $0.06$& $71.5$ & $0.03$ & $76.5$ & $0.03$ & $71.7$ & $0.03$ & $76.4$ & $0.03$ & $71.6$ & $0.03$ & $75.7$ & $0.03$\\ 
& $0$ & $0$ & $1$ & $1$ & $78.6$ & $0.06$ & $82.4$ & $0.04$ & $78.8$ & $0.05$ & $83.3$ & $0.04$ & $78.7$ & $0.05$ & $82.6$ & $0.04$& $74.4$ & $0.05$ & $77.4$ & $0.05$ & $74.6$ & $0.05$ & $77.3$ & $0.04$ & $74.5$ & $0.05$ & $76.6$ & $0.04$\\ 
\hline
\hline
\multirow{3}{*}{R}
& $0$ & $0$ & $0$ & $0$ & $80.3$ & $0.08$ & $84.2$ & $0.07$ & $80.5$ & $0.07$ & $85.1$ & $0.06$ & $80.4$ & $0.08$ & $84.4$ & $0.07$& $80.2$ & $0.09$ & $84.2$ & $0.08$ & $80.4$ & $0.08$ & $85.1$ & $0.07$ & $80.3$ & $0.09$ & $84.4$ & $0.08$\\ 
& $0$ & $0$ & $1$ & $0$ & $75.3$ & $0.02$ & $82.6$ & $0.01$ & $75.5$ & $0.02$ & $83.5$ & $0.01$ & $75.4$ & $0.02$ & $82.8$ & $0.01$& $75.5$ & $0.04$ & $82.4$ & $0.03$ & $75.7$ & $0.04$ & $83.3$ & $0.03$ & $75.6$ & $0.04$ & $82.6$ & $0.03$\\ 
& $0$ & $0$ & $1$ & $1$ & $78.4$ & $0.03$ & $83.4$ & $0.03$ & $78.6$ & $0.03$ & $84.3$ & $0.02$ & $78.5$ & $0.03$ & $83.6$ & $0.03$& $78.5$ & $0.05$ & $83.5$ & $0.02$ & $78.7$ & $0.04$ & $84.4$ & $0.02$ & $78.6$ & $0.05$ & $83.7$ & $0.02$\\ 
\hline
\hline
\multirow{3}{*}{\tiny G+R}
& $0$ & $0$ & $0$ & $0$ & $80.2$ & $0.16$ & $84.6$ & $0.14$ & $80.4$ & $0.14$ & $85.3$ & $0.14$ & $80.3$ & $0.15$ & $84.9$ & $0.14$& $80.2$ & $0.16$ & $84.8$ & $0.14$ & $80.4$ & $0.14$ & $85.5$ & $0.14$ & $80.3$ & $0.15$ & $85.1$ & $0.14$\\ 
& $0$ & $0$ & $1$ & $0$ & $75.2$ & $0.05$ & $83.2$ & $0.04$ & $75.3$ & $0.04$ & $83.9$ & $0.04$ & $75.3$ & $0.05$ & $83.5$ & $0.04$& $75.3$ & $0.05$ & $83.1$ & $0.05$ & $75.5$ & $0.04$ & $83.8$ & $0.05$ & $75.4$ & $0.05$ & $83.4$ & $0.05$\\ 
& $0$ & $0$ & $1$ & $1$ & $78.5$ & $0.05$ & $83.9$ & $0.05$ & $78.7$ & $0.04$ & $84.6$ & $0.05$ & $78.6$ & $0.05$ & $84.2$ & $0.05$& $78.6$ & $0.06$ & $84.1$ & $0.04$ & $78.8$ & $0.05$ & $84.7$ & $0.04$ & $78.7$ & $0.06$ & $84.3$ & $0.04$\\ 
\hline
\hline
\end{tabular}
\caption{Results for a shared model trained with STL and MTL, with or without the sensitive feature as predictor, and with or without the fairness constraint.}
\label{tab:ris1}
\end{table*}
From Table~\ref{tab:ris1} it is possible to see that MTL reaches higher accuracies compared to STL while the fairness measure is mostly comparable, this means that there is a relation between the tasks which can be captured with MTL. 
%\LO{This hypothesis is also supported by the results of Figure~\ref{fig:tableexplanation} where we fix $\theta$ and $\rho$, to be the best values as observed in MTL, which we call respectively $\theta^*$ and $\rho^*$
This hypothesis is also supported by the results of Figure~\ref{fig:tableexplanation}, in which we 
%which we fix $\theta$ and $\rho$ in Eq.~\eqref{eq:MTL} to be the best values found during the CV procedure
%that gave us the best results as observed in MTL, which we call respectively $\theta^*$ and $\rho^*$, 
check how the accuracy and fairness, as measured with the EOd, varies by varying $\lambda$.
Figure~\ref{fig:tableexplanation} shows that there are commonalities
%correlations
between the groups which increase by increasing the number of groups: 
the optimal parameter $\lambda$ 
%\AM{Don't really get this sentence. The correlation increases between groups 
%if the total number of groups is larger?} 
%(\AM{Suggested rephrase: When D=0, we have that $\lambda <1$, and we consider the shared model. 
%When $D=1$, we find that $\lambda > 0$, and we consider the group specific models}
it is smaller than one when we consider the shared model ($D{=}0$) and it is larger than zeros 
%\AM{isn't $\lambda$ also larger than 0 for $\lambda <1$} 
when we consider group specific models ($D{=}1$).
Moreover, as expected the fairness constraint has a negative impact on the accuracy (less strong for MTL) whilst having a highly positive impact on fairness.
Having the sensitive feature as a predictor increases the accuracy, but decreases the fairness measure, as expected.

Table~\ref{tab:ris2} reports the case when the group specific models are trained with ITL or MTL, the same setting as Table~\ref{tab:ris1}.
MTL {notably} improves both accuracy and fairness.
The fairness constraints do not affect the accuracy too much, while giving remarkable improvements in fairness.
ITL and MTL are not affected by not including or including the sensitive feature predictor, as expected from the theory given that the models already have already different biases.
\begin{table*}[!h]
\footnotesize
\centering
\setlength{\tabcolsep}{0.04cm}
\renewcommand{\arraystretch}{1.}
\begin{tabular}{||c||c|c|c|c||||c|c||c|c|||c|c||c|c|||c|c||c|c||||c|c||c|c|||c|c||c|c|||c|c||c|c||||}
\hline
\hline
\multicolumn{5}{||c||||}{} 
& \multicolumn{12}{c||||}{Adult Dataset}
& \multicolumn{12}{c||||}{COMPAS Dataset} \\
\cline{2-29}
& $-$ & $0$ & $-$ & $-$ 
& \multicolumn{2}{c||}{STL} & \multicolumn{2}{c|||}{\multirow{2}{*}{MTL}} 
& \multicolumn{2}{c||}{STL} & \multicolumn{2}{c|||}{\multirow{2}{*}{MTL}} 
& \multicolumn{2}{c||}{STL} & \multicolumn{2}{c||||}{\multirow{2}{*}{MTL}}
& \multicolumn{2}{c||}{STL} & \multicolumn{2}{c|||}{\multirow{2}{*}{MTL}} 
& \multicolumn{2}{c||}{STL} & \multicolumn{2}{c|||}{\multirow{2}{*}{MTL}} 
& \multicolumn{2}{c||}{STL} & \multicolumn{2}{c||||}{\multirow{2}{*}{MTL}}\\
& $-$ & $1$ & $-$ & $-$ 
& \multicolumn{2}{c||}{ITL} & \multicolumn{2}{c|||}{} 
& \multicolumn{2}{c||}{ITL} & \multicolumn{2}{c|||}{} 
& \multicolumn{2}{c||}{ITL} & \multicolumn{2}{c||||}{}
& \multicolumn{2}{c||}{ITL} & \multicolumn{2}{c|||}{} 
& \multicolumn{2}{c||}{ITL} & \multicolumn{2}{c|||}{} 
& \multicolumn{2}{c||}{ITL} & \multicolumn{2}{c||||}{}\\
\cline{2-29}
& P & D & F & S 
& \tiny ACC & \tiny DEOp$^+$ & \tiny ACC & \tiny DEOp$^+$ 
& \tiny ACC & \tiny DEOp$^-$ & \tiny ACC & \tiny DEOp$^-$ 
& \tiny \tiny ACC & \tiny DEOd & \tiny ACC & \tiny DEOd 
& \tiny ACC & \tiny DEOp$^+$ & \tiny ACC & \tiny DEOp$^+$ 
& \tiny ACC & \tiny DEOp$^-$ & \tiny ACC & \tiny DEOp$^-$ 
& \tiny \tiny ACC & \tiny DEOd & \tiny ACC & \tiny DEOd \\ 
\hline
\hline
\multirow{3}{*}{G}
& $0$ & $1$ & $0$ & $0$ & $74.5$ & $0.18$ & $90.0$ & $0.14$ & $74.7$ & $0.15$ & $91.0$ & $0.13$ & $74.6$ & $0.17$ & $90.2$ & $0.14$& $70.7$ & $0.19$ & $84.5$ & $0.15$ & $70.9$ & $0.17$ & $84.4$ & $0.14$ & $70.8$ & $0.16$ & $83.6$ & $0.13$\\ 
& $0$ & $1$ & $1$ & $0$ & $69.7$ & $0.08$ & $88.3$ & $0.04$ & $69.9$ & $0.07$ & $89.2$ & $0.04$ & $69.8$ & $0.08$ & $88.5$ & $0.04$& $66.1$ & $0.08$ & $83.0$ & $0.04$ & $66.3$ & $0.08$ & $82.8$ & $0.04$ & $66.2$ & $0.07$ & $82.1$ & $0.04$\\ 
& $0$ & $1$ & $1$ & $1$ & $69.7$ & $0.08$ & $88.1$ & $0.03$ & $69.9$ & $0.07$ & $89.1$ & $0.03$ & $69.8$ & $0.08$ & $88.3$ & $0.03$& $66.1$ & $0.09$ & $82.9$ & $0.07$ & $66.3$ & $0.08$ & $82.8$ & $0.06$ & $66.2$ & $0.08$ & $82.1$ & $0.06$\\ 
\hline
\hline
\multirow{3}{*}{R}
& $0$ & $1$ & $0$ & $0$ & $67.4$ & $0.13$ & $91.8$ & $0.10$ & $67.6$ & $0.11$ & $92.8$ & $0.08$ & $67.5$ & $0.13$ & $92.0$ & $0.10$& $67.3$ & $0.12$ & $91.7$ & $0.08$ & $67.5$ & $0.11$ & $92.7$ & $0.07$ & $67.4$ & $0.12$ & $92.0$ & $0.08$\\ 
& $0$ & $1$ & $1$ & $0$ & $62.5$ & $0.05$ & $90.0$ & $0.03$ & $62.7$ & $0.05$ & $90.9$ & $0.03$ & $62.6$ & $0.05$ & $90.2$ & $0.03$& $62.4$ & $0.07$ & $90.1$ & $0.02$ & $62.6$ & $0.06$ & $91.0$ & $0.02$ & $62.5$ & $0.07$ & $90.3$ & $0.02$\\ 
& $0$ & $1$ & $1$ & $1$ & $62.6$ & $0.06$ & $90.4$ & $0.03$ & $62.7$ & $0.05$ & $91.3$ & $0.03$ & $62.6$ & $0.06$ & $90.6$ & $0.03$& $62.4$ & $0.07$ & $90.0$ & $0.03$ & $62.5$ & $0.07$ & $91.0$ & $0.03$ & $62.4$ & $0.07$ & $90.2$ & $0.03$\\ 
\hline
\hline
\multirow{3}{*}{\tiny G+R}
& $0$ & $1$ & $0$ & $0$ & $64.0$ & $0.23$ & $91.5$ & $0.15$ & $64.2$ & $0.20$ & $92.2$ & $0.15$ & $64.1$ & $0.22$ & $91.8$ & $0.15$& $64.2$ & $0.24$ & $91.4$ & $0.16$ & $64.3$ & $0.21$ & $92.2$ & $0.16$ & $64.3$ & $0.22$ & $91.7$ & $0.16$\\
& $0$ & $1$ & $1$ & $0$ & $59.3$ & $0.14$ & $89.8$ & $0.05$ & $59.4$ & $0.12$ & $90.6$ & $0.05$ & $59.3$ & $0.13$ & $90.1$ & $0.05$& $59.2$ & $0.13$ & $90.1$ & $0.05$ & $59.4$ & $0.11$ & $90.8$ & $0.05$ & $59.3$ & $0.12$ & $90.4$ & $0.05$\\ 
& $0$ & $1$ & $1$ & $1$ & $59.2$ & $0.13$ & $90.0$ & $0.05$ & $59.4$ & $0.11$ & $90.8$ & $0.05$ & $59.3$ & $0.12$ & $90.3$ & $0.05$& $59.4$ & $0.13$ & $89.9$ & $0.05$ & $59.5$ & $0.11$ & $90.7$ & $0.05$ & $59.5$ & $0.12$ & $90.3$ & $0.05$\\ 
\hline
\hline
\end{tabular}
\caption{Results when group specific models are trained with ITL and MTL with or without the sensitive feature as predictor and with or without the fairness constraint.}
\label{tab:ris2}
\end{table*}
Table~\ref{tab:STL-ITL-MTL-minority} reports a comparison between STL, ITL, and MTL on the Adult dataset, showing the accuracy on each group for the different models for the case that $P{=}0$, $F{=}0$, and $S{=}0$. These results clearly demonstrate that STL and ITL tend to generalize poorly on smaller groups, whereas MTL generalizes better.
Results on COMPAS datasets are analogous.
\begin{table}
\centering
\setlength{\tabcolsep}{0.08cm}
\renewcommand{\arraystretch}{1}
\begin{tabular}{|c|l||c|c||c|c||}
\hline
\hline
 & & \multicolumn{2}{c||}{D${=}0$} & \multicolumn{2}{c||}{D${=}1$} \\
Sens. & Group & STL & MTL & ITL & MTL \\
\hline
\hline
\multirow{2}{*}{G} 
& M & 85.4 & 88.5 & 78.8 & 92.8 \\
& F & 81.2 & 85.9 & 74.2 & 91.0 \\ 
\hline
\multirow{5}{*}{R} 
& W & 86.7 & 89.8 & 89.7 & 93.2 \\
& B & 83.5 & 88.9 & 83.5 & 92.8 \\
& API & 82.3 & 87.9 & 65.2 & 92.1 \\
& AIE & 82.1 & 87.6 & 48.5 & 92.0 \\
& O & 81.2 & 86.9 & 47.5 & 92.1 \\
\hline
\multirow{10}{*}{\tiny G+R}
& W\&M & 87.8 & 92.8 & 85.8 & 94.7 \\
& W\&F & 85.6 & 89.5 & 84.7 & 93.2 \\
& B\&M & 84.4 & 89.9 & 66.3 & 93.2 \\
& B\&F & 82.4 & 88.1 & 64.6 & 92.1 \\
& API\&M & 83.6 & 89.2 & 67.3 & 93.0 \\
& API\&F & 81.8 & 88.0 & 63.5 & 92.8 \\
& AIE\&M & 83.0 & 88.8 & 50.2 & 92.7 \\
& AIE\&F & 81.9 & 87.3 & 45.1 & 92.5 \\
& O\&M & 81.7 & 88.3 & 50.7 & 93.1 \\
& O\&F & 81.1 & 86.6 & 43.3 & 92.1 \\
\hline
\hline
\end{tabular}
\caption{Adult dataset: ACC of STL, ITL, and MTL when $P{=}0$, $F{=}0$, and $S{=}0$.}
\label{tab:STL-ITL-MTL-minority}
\end{table}

Table~\ref{tab:ris3} reports the comparison between the most accurate, fair and legal\footnote{From now, for sake of simplicity, we use the word illegal (legal) in order to define a model which uses (not-uses), either implicitly or explicitly, the sensitive feature as part of its functional form.} model (the shared model trained with MTL, with fairness constraint, and no sensitive feature in the predictors) and the most accurate, fair and illegal model (the group specific models trained with MTL, with fairness constraint, the sensitive feature used as predictor).
From the table one can note that the illegal model remarkably improves over the legal one in terms of accuracy and in some cases it is even better than the legal one in terms of fairness.
\begin{table}
\centering
\setlength{\tabcolsep}{0.08cm}
\renewcommand{\arraystretch}{1}
\begin{tabular}{||c||c|c|c|c||c|c||c|c||c|c||c|c||c|c||c|c||||}
\hline
\hline
& \multicolumn{4}{c||}{} & \multicolumn{2}{c||}{MTL} & \multicolumn{2}{c||}{MTL} & \multicolumn{2}{c||}{MTL} \\
& P & D & F & S 
& \tiny ACC & \tiny DEOp$^+$ 
& \tiny ACC & \tiny DEOp$^-$ 
& \tiny ACC & \tiny DEOd \\ 
\hline
\hline
\multicolumn{11}{c}{Adult Dataset}\\
\hline
\hline
\multirow{2}{*}{G}
& $0$ & $0$ & $1$ & $0$ & $81.8$ & $0.06$ & $82.7$ & $0.05$ & $82.0$ & $0.06$\\ 
& $0$ & $1$ & $1$ & $1$ & $88.1$ & $0.03$ & $89.1$ & $0.03$ & $88.3$ & $0.03$\\ 
\hline
\hline
\multirow{2}{*}{R}
& $0$ & $0$ & $1$ & $0$ & $82.6$ & $0.01$ & $83.5$ & $0.01$ & $82.8$ & $0.01$\\ 
& $0$ & $1$ & $1$ & $1$ & $90.4$ & $0.03$ & $91.3$ & $0.03$ & $90.6$ & $0.03$\\ 
\hline
\hline
\multirow{2}{*}{\tiny G+R}
& $0$ & $0$ & $1$ & $0$ & $83.2$ & $0.04$ & $83.9$ & $0.04$ & $83.5$ & $0.04$\\ 
& $0$ & $1$ & $1$ & $1$ & $90.0$ & $0.05$ & $90.8$ & $0.05$ & $90.3$ & $0.05$\\ 
\hline
\hline
\multicolumn{11}{c}{COMPAS Dataset}\\
\hline
\hline
\multirow{2}{*}{G}
& $0$ & $0$ & $1$ & $0$ & $76.5$ & $0.03$ & $76.4$ & $0.03$ & $75.7$ & $0.03$\\ 
& $0$ & $1$ & $1$ & $1$ & $82.9$ & $0.07$ & $82.8$ & $0.06$ & $82.1$ & $0.06$\\ 
\hline
\hline
\multirow{2}{*}{R}
& $0$ & $0$ & $1$ & $0$ & $82.4$ & $0.03$ & $83.3$ & $0.03$ & $82.6$ & $0.03$\\ 
& $0$ & $1$ & $1$ & $1$ & $90.0$ & $0.03$ & $91.0$ & $0.03$ & $90.2$ & $0.03$\\ 
\hline
\hline
\multirow{2}{*}{\tiny G+R}
& $0$ & $0$ & $1$ & $0$ & $83.1$ & $0.05$ & $83.8$ & $0.05$ & $83.4$ & $0.05$\\ 
& $0$ & $1$ & $1$ & $1$ & $89.9$ & $0.05$ & $90.7$ & $0.05$ & $90.3$ & $0.05$\\ 
\hline
\hline
\end{tabular}
\caption{Comparison between the most accurate, fair and legal model (the shared model trained with MTL, with fairness constraint, and no sensitive feature in the predictors) and the most accurate, fair and illegal model (the group specific models trained with MTL, with fairness constraint, the sensitive feature exploited as predictor).}
\label{tab:ris3}
\end{table}
Based on the result of Table~\ref{tab:ris3} we would like to be able to use the 'illegal' model'. In order to do so make use of the trick described in the previous sections, namely we use the predicted sensitive feature based on the non-sensitive features, instead of the true one. For this purpose we used a Random Forests model~\cite{breiman2001random} where we weighted the errors differently based on the group membership.
Table~\ref{tab:confusion} and Table~\ref{tab:COMPASconfusion} report the confusion matrices computed on the test set.
%MMM of the test error on the test set.

Finally, in Table~\ref{tab:ris4} we report a comparison between the best illegal model and the same model, but for which uses we used the predicted sensitive feature, instead of the true one, both in training and in testing. 
Notably, Table~\ref{tab:ris4} shows that using the predicted sensitive feature in place of the true one 
preserves the accuracy of the learned model, but with a 
%dramatic
notable improvement in fairness.
In attempt to explain this phenomena, in Table~\ref{tab:distanceanderror} we report the average 
group accuracy for predicting the sensitive features gender and race, as a function of the distance 
from the group specific models separators trained with MTL on the Adult dataset. 
%\AM{as a function of the distance of the group specific model, to the separators trained with MTL? Don't really understand this sentence.}. 
Table~\ref{tab:distanceanderror} shows that the accuracy in predicting the sensitive feature decreases as we 
get closer to the separator. This can be understood as allowing the group specific model to randomize which specific classifier to use, reducing overall unfairness of the decision.
Results on COMPAS dataset are analogous.
%
%\LO{<I think that this is not important>}
%\MP{<If we use the predicted feature with ITL, do we do the same or (much) worse than MTL with predicted feature?>}

\begin{table}
\centering
\setlength{\tabcolsep}{0.08cm}
\renewcommand{\arraystretch}{1}
\begin{tabular}{||c||c|c|c|c||c|c||c|c||c|c||c|c||c|c||c|c||||}
\hline
\hline
& \multicolumn{4}{c||}{} & \multicolumn{2}{c||}{MTL} & \multicolumn{2}{c||}{MTL} & \multicolumn{2}{c||}{MTL} \\
& P & D & F & S 
& \tiny ACC & \tiny DEOp$^+$ 
& \tiny ACC & \tiny DEOp$^-$ 
& \tiny ACC & \tiny DEOd \\ 
\hline
\hline
\multicolumn{11}{c}{Adult Dataset}\\
\hline
\hline
\multirow{2}{*}{G}
& $0$ & $1$ & $1$ & $1$ & $88.1$ & $0.03$ & $89.1$ & $0.03$ & $88.3$ & $0.03$\\ 
& $1$ & $1$ & $1$ & $1$ & $87.4$ & $0.01$ & $88.3$ & $0.01$ & $87.6$ & $0.01$\\ 
\hline
\hline
\multirow{2}{*}{R}
& $0$ & $1$ & $1$ & $1$ & $90.4$ & $0.03$ & $91.3$ & $0.03$ & $90.6$ & $0.03$\\ 
& $1$ & $1$ & $1$ & $1$ & $89.2$ & $0.01$ & $90.2$ & $0.01$ & $89.4$ & $0.01$\\ 
\hline
\hline
\multirow{2}{*}{\tiny G+R}
& $0$ & $1$ & $1$ & $1$ & $90.0$ & $0.05$ & $90.8$ & $0.05$ & $90.3$ & $0.05$\\ 
& $1$ & $1$ & $1$ & $1$ & $89.0$ & $0.01$ & $89.8$ & $0.01$ & $89.3$ & $0.01$\\ 
\hline
\hline
\multicolumn{11}{c}{COMPAS Dataset}\\
\hline
\hline
\multirow{2}{*}{G}
& $0$ & $1$ & $1$ & $1$ & $82.9$ & $0.07$ & $82.8$ & $0.06$ & $82.1$ & $0.06$\\ 
& $1$ & $1$ & $1$ & $1$ & $82.1$ & $0.01$ & $82.0$ & $0.01$ & $81.3$ & $0.01$\\ 
\hline
\hline
\multirow{2}{*}{R}
& $0$ & $1$ & $1$ & $1$ & $90.0$ & $0.03$ & $91.0$ & $0.03$ & $90.2$ & $0.03$\\ 
& $1$ & $1$ & $1$ & $1$ & $89.0$ & $0.01$ & $89.9$ & $0.01$ & $89.2$ & $0.01$\\ 
\hline
\hline
\multirow{2}{*}{\tiny G+R}
& $0$ & $1$ & $1$ & $1$ & $89.9$ & $0.05$ & $90.7$ & $0.05$ & $90.3$ & $0.05$\\ 
& $1$ & $1$ & $1$ & $1$ & $89.0$ & $0.01$ & $89.8$ & $0.01$ & $89.3$ & $0.01$\\ 
\hline
\hline
\end{tabular}
\caption{Comparison between the group specific models trained with MTL, with fairness constraint, and the true sensitive feature exploited as a predictor, against the same model when the predicted sensitive feature exploited as predictor.
%\AM{comparison between the group specific models trained with MTL, with fairness constraints, as well as using the true sensitive feature as a predictor, versus the same model, but with the predicted sensitive feature as a predictor instead.}
}
\label{tab:ris4}
\end{table}
\begin{table}
\centering
\setlength{\tabcolsep}{0.08cm}
\renewcommand{\arraystretch}{1}
\begin{tabular}{|c|c|c|c|c|c|c|}
\hline
\hline
& \multicolumn{3}{c|}{Margin Distance} \\
 & $1/10$ & $1$ & $\infty$ \\
\hline
\hline
G & $75.4$ & $ 83.3 $ & $87.3$ \\
\hline
R & $69.9$ & $ 80.7 $ & $84.7$ \\
\hline
\hline
\end{tabular}
\caption{Adult dataset: accuracy in percentage of prediction based on the distance from the MTL separator which uses the predicted sensitive feature (see Table~\ref{tab:ris4}).}
\label{tab:distanceanderror}
\end{table}
%
%
% Do not repeat the list and refere to to Section 3 or Section 1
 
%\vspace{-.2cm}
\section{Discussion}
\label{sec:Discussion}
%\vspace{-.15cm}
%
%\AM{We have studied a multitask learning framework enhanced with fairness constraints. Multitask learning naturally deals with the issue of having little data on minority groups. However, often certain measures of fairness need to be specifically enforced. Popular measures are the EOp and EOd, and we have shown how to enhance multitask learning with these fairness constraints. Furthermore, there exists an unavoidable trade-off between accuracy gains obtained by multitask learning, which necessitates the use of sensitive features, and any commitment to protect these characteristics. Therefore, to address this issue, we have proposed to use predicted sensitive features instead. In this way we have developed a method with increased accuracy and fairness, which treats different individuals equally.}%
We have presented two novel, but related, ideas in this work.
Firstly, to resolve the tension between accuracy gains obtained by using a sensitive feature as part of the model, and the potential inapplicability of such an approach, 
we have suggested to first predict the sensitive feature based on the non-sensitive features, and then use the predicted value in the functional form of a model, allowing to treat people belonging to different groups, but having similar non-sensitive features, equally.
%us to overcome any legal issues.
Furthermore, we have demonstrated how the predicted sensitive feature can then used in a fairness constrained MTL framework.
We confirmed the validity of the above approach empirically, giving us substantial improvements in both accuracy and fairness, compared to STL and ITL. 
We believe this to be a fruitful area of possible future research.
Of course, a non-linear extension of the above framework would be interesting to study, although we did not notice any substantial improvements on the Adult and COMPAS datasets considered in this work. Moreover, it would be interesting to see if the above framework can be extended to include other fairness definitions, apart from the EOp and EOd that we have tested.
%Another interesting direction of research is to investigate well-known methods from MTL, such as those based on feature learning \cite{argyriou2008convex,caruana1997multitask}, and apply these methods in the context of fairness.
%Theoretically, 
Finally, it would be valuable to provide theoretical conditions on the data distribution for which our approach provably works. %studying what type of 
%group structure lends itself to achieve the type of remarkable results that we have demonstrated on the Adult dataset.

\begin{acks} 
This work was supported by the Amazon AWS Machine Learning Research Award.
\end{acks}

\bibliographystyle{ACM-Reference-Format}
\bibliography{biblio}


\begin{thebibliography}{39}

%%% the Web address.

\ifx \showCODEN    \undefined \def \showCODEN     #1{\unskip}     \fi
\ifx \showDOI      \undefined \def \showDOI       #1{#1}\fi
\ifx \showISBNx    \undefined \def \showISBNx     #1{\unskip}     \fi
\ifx \showISBNxiii \undefined \def \showISBNxiii  #1{\unskip}     \fi
\ifx \showISSN     \undefined \def \showISSN      #1{\unskip}     \fi
\ifx \showLCCN     \undefined \def \showLCCN      #1{\unskip}     \fi
\ifx \shownote     \undefined \def \shownote      #1{#1}          \fi
\ifx \showarticletitle \undefined \def \showarticletitle #1{#1}   \fi
\ifx \showURL      \undefined \def \showURL       {\relax}        \fi
% invisible to TeX
\providecommand\bibfield[2]{#2}
\providecommand\bibinfo[2]{#2}
\providecommand\natexlab[1]{#1}
\providecommand\showeprint[2][]{arXiv:#2}

\bibitem[\protect\citeauthoryear{Adebayo and Kagal}{Adebayo and Kagal}{2016}]%
        {adebayo2016iterative}
\bibfield{author}{\bibinfo{person}{J. Adebayo} {and} \bibinfo{person}{L.
  Kagal}.} \bibinfo{year}{2016}\natexlab{}.
\newblock \showarticletitle{Iterative orthogonal feature projection for
  diagnosing bias in black-box models}. In \bibinfo{booktitle}{\emph{Conference
  on Fairness, Accountability, and Transparency in Machine Learning}}.
\newblock


\bibitem[\protect\citeauthoryear{Agarwal, Beygelzimer, Dud{\'\i}k, and
  Langford}{Agarwal et~al\mbox{.}}{2017}]%
        {agarwal2017reductions}
\bibfield{author}{\bibinfo{person}{A. Agarwal}, \bibinfo{person}{A.
  Beygelzimer}, \bibinfo{person}{M. Dud{\'\i}k}, {and} \bibinfo{person}{J.
  Langford}.} \bibinfo{year}{2017}\natexlab{}.
\newblock \showarticletitle{A Reductions Approach to Fair Classification}. In
  \bibinfo{booktitle}{\emph{Conference on Fairness, Accountability, and
  Transparency in Machine Learning}}.
\newblock


\bibitem[\protect\citeauthoryear{Agarwal, Beygelzimer, Dud{\'\i}k, Langford,
  and Wallach}{Agarwal et~al\mbox{.}}{2018}]%
        {agarwal2018reductions}
\bibfield{author}{\bibinfo{person}{A. Agarwal}, \bibinfo{person}{A.
  Beygelzimer}, \bibinfo{person}{M. Dud{\'\i}k}, \bibinfo{person}{J. Langford},
  {and} \bibinfo{person}{H. Wallach}.} \bibinfo{year}{2018}\natexlab{}.
\newblock \showarticletitle{A reductions approach to fair classification}.
\newblock \bibinfo{journal}{\emph{arXiv preprint arXiv:1803.02453}}
  (\bibinfo{year}{2018}).
\newblock


\bibitem[\protect\citeauthoryear{Alabi, Immorlica, and Kalai}{Alabi
  et~al\mbox{.}}{2018}]%
        {alabi2018optimizing}
\bibfield{author}{\bibinfo{person}{D. Alabi}, \bibinfo{person}{N. Immorlica},
  {and} \bibinfo{person}{A.~T. Kalai}.} \bibinfo{year}{2018}\natexlab{}.
\newblock \showarticletitle{When optimizing nonlinear objectives is no harder
  than linear objectives}.
\newblock \bibinfo{journal}{\emph{arXiv preprint arXiv:1804.04503}}
  (\bibinfo{year}{2018}).
\newblock


\bibitem[\protect\citeauthoryear{Argyriou, Evgeniou, and Pontil}{Argyriou
  et~al\mbox{.}}{2008}]%
        {argyriou2008convex}
\bibfield{author}{\bibinfo{person}{A. Argyriou}, \bibinfo{person}{T. Evgeniou},
  {and} \bibinfo{person}{M. Pontil}.} \bibinfo{year}{2008}\natexlab{}.
\newblock \showarticletitle{Convex multi-task feature learning}.
\newblock \bibinfo{journal}{\emph{Machine Learning}} \bibinfo{volume}{73},
  \bibinfo{number}{3} (\bibinfo{year}{2008}), \bibinfo{pages}{243--272}.
\newblock


\bibitem[\protect\citeauthoryear{Bakker and Heskes}{Bakker and Heskes}{2003}]%
        {bakker2003task}
\bibfield{author}{\bibinfo{person}{B. Bakker} {and} \bibinfo{person}{T.
  Heskes}.} \bibinfo{year}{2003}\natexlab{}.
\newblock \showarticletitle{Task clustering and gating for bayesian multitask
  learning}.
\newblock \bibinfo{journal}{\emph{Journal of Machine Learning Research}}
  \bibinfo{volume}{4} (\bibinfo{year}{2003}), \bibinfo{pages}{83--99}.
\newblock


\bibitem[\protect\citeauthoryear{Baxter}{Baxter}{2000}]%
        {baxter2000model}
\bibfield{author}{\bibinfo{person}{J. Baxter}.}
  \bibinfo{year}{2000}\natexlab{}.
\newblock \showarticletitle{A model of inductive bias learning}.
\newblock \bibinfo{journal}{\emph{Journal of artificial intelligence research}}
   \bibinfo{volume}{12} (\bibinfo{year}{2000}), \bibinfo{pages}{149--198}.
\newblock


\bibitem[\protect\citeauthoryear{Bechavod and Ligett}{Bechavod and
  Ligett}{2018}]%
        {bechavod2018Penalizing}
\bibfield{author}{\bibinfo{person}{Y. Bechavod} {and} \bibinfo{person}{K.
  Ligett}.} \bibinfo{year}{2018}\natexlab{}.
\newblock \showarticletitle{Penalizing Unfairness in Binary Classification}.
\newblock \bibinfo{journal}{\emph{arXiv preprint arXiv:1707.00044v3}}
  (\bibinfo{year}{2018}).
\newblock


\bibitem[\protect\citeauthoryear{Berk, Heidari, Jabbari, Joseph, Kearns,
  Morgenstern, Neel, and Roth}{Berk et~al\mbox{.}}{2017}]%
        {berk2017convex}
\bibfield{author}{\bibinfo{person}{R. Berk}, \bibinfo{person}{H. Heidari},
  \bibinfo{person}{S. Jabbari}, \bibinfo{person}{M. Joseph},
  \bibinfo{person}{M. Kearns}, \bibinfo{person}{J. Morgenstern},
  \bibinfo{person}{S. Neel}, {and} \bibinfo{person}{A. Roth}.}
  \bibinfo{year}{2017}\natexlab{}.
\newblock \showarticletitle{A convex framework for fair regression}.
\newblock \bibinfo{journal}{\emph{arXiv preprint arXiv:1706.02409}}
  (\bibinfo{year}{2017}).
\newblock


\bibitem[\protect\citeauthoryear{Beutel, Chen, Zhao, and Chi}{Beutel
  et~al\mbox{.}}{2017}]%
        {beutel2017data}
\bibfield{author}{\bibinfo{person}{A. Beutel}, \bibinfo{person}{J. Chen},
  \bibinfo{person}{Z. Zhao}, {and} \bibinfo{person}{E.~H. Chi}.}
  \bibinfo{year}{2017}\natexlab{}.
\newblock \showarticletitle{Data decisions and theoretical implications when
  adversarially learning fair representations}. In
  \bibinfo{booktitle}{\emph{Conference on Fairness, Accountability, and
  Transparency in Machine Learning}}.
\newblock


\bibitem[\protect\citeauthoryear{Breiman}{Breiman}{2001}]%
        {breiman2001random}
\bibfield{author}{\bibinfo{person}{L. Breiman}.}
  \bibinfo{year}{2001}\natexlab{}.
\newblock \showarticletitle{Random forests}.
\newblock \bibinfo{journal}{\emph{Machine learning}} \bibinfo{volume}{45},
  \bibinfo{number}{1} (\bibinfo{year}{2001}), \bibinfo{pages}{5--32}.
\newblock


\bibitem[\protect\citeauthoryear{Calmon, Wei, Vinzamuri, Ramamurthy, and
  Varshney}{Calmon et~al\mbox{.}}{2017}]%
        {calmon2017optimized}
\bibfield{author}{\bibinfo{person}{F. Calmon}, \bibinfo{person}{D. Wei},
  \bibinfo{person}{B. Vinzamuri}, \bibinfo{person}{K.~Natesan Ramamurthy},
  {and} \bibinfo{person}{K.~R. Varshney}.} \bibinfo{year}{2017}\natexlab{}.
\newblock \showarticletitle{Optimized Pre-Processing for Discrimination
  Prevention}. In \bibinfo{booktitle}{\emph{Advances in Neural Information
  Processing Systems}}.
\newblock


\bibitem[\protect\citeauthoryear{Caruana}{Caruana}{1997}]%
        {caruana1997multitask}
\bibfield{author}{\bibinfo{person}{R. Caruana}.}
  \bibinfo{year}{1997}\natexlab{}.
\newblock \showarticletitle{Multitask Learning}.
\newblock \bibinfo{journal}{\emph{Machine Learning}} \bibinfo{volume}{28},
  \bibinfo{number}{1} (\bibinfo{year}{1997}), \bibinfo{pages}{41--75}.
\newblock


\bibitem[\protect\citeauthoryear{Chouldechova}{Chouldechova}{2017}]%
        {chouldechova2017fair}
\bibfield{author}{\bibinfo{person}{A. Chouldechova}.}
  \bibinfo{year}{2017}\natexlab{}.
\newblock \showarticletitle{Fair prediction with disparate impact: A study of
  bias in recidivism prediction instruments}.
\newblock \bibinfo{journal}{\emph{Big data}} \bibinfo{volume}{5},
  \bibinfo{number}{2} (\bibinfo{year}{2017}), \bibinfo{pages}{153--163}.
\newblock


\bibitem[\protect\citeauthoryear{Donini, Martinez-Rego, Goodson, Shawe-Taylor,
  and Pontil}{Donini et~al\mbox{.}}{2016}]%
        {donini2016distributed}
\bibfield{author}{\bibinfo{person}{Michele Donini}, \bibinfo{person}{David
  Martinez-Rego}, \bibinfo{person}{Martin Goodson}, \bibinfo{person}{John
  Shawe-Taylor}, {and} \bibinfo{person}{Massimiliano Pontil}.}
  \bibinfo{year}{2016}\natexlab{}.
\newblock \showarticletitle{Distributed variance regularized multitask
  learning}. In \bibinfo{booktitle}{\emph{Neural Networks (IJCNN), 2016
  International Joint Conference on}}. IEEE, \bibinfo{pages}{3101--3109}.
\newblock


\bibitem[\protect\citeauthoryear{Donini, Oneto, Ben-David, Shawe-Taylor, and
  Pontil}{Donini et~al\mbox{.}}{2018}]%
        {donini2018empirical}
\bibfield{author}{\bibinfo{person}{M. Donini}, \bibinfo{person}{L. Oneto},
  \bibinfo{person}{S. Ben-David}, \bibinfo{person}{J. Shawe-Taylor}, {and}
  \bibinfo{person}{M. Pontil}.} \bibinfo{year}{2018}\natexlab{}.
\newblock \showarticletitle{Empirical Risk Minimization under Fairness
  Constraints}.
\newblock \bibinfo{journal}{\emph{arXiv preprint arXiv:1802.08626}}
  (\bibinfo{year}{2018}).
\newblock


\bibitem[\protect\citeauthoryear{Dwork, Immorlica, Kalai, and Leiserson}{Dwork
  et~al\mbox{.}}{2018}]%
        {dwork2018decoupled}
\bibfield{author}{\bibinfo{person}{C. Dwork}, \bibinfo{person}{N. Immorlica},
  \bibinfo{person}{A.~T. Kalai}, {and} \bibinfo{person}{M.~D.~M. Leiserson}.}
  \bibinfo{year}{2018}\natexlab{}.
\newblock \showarticletitle{Decoupled Classifiers for Group-Fair and Efficient
  Machine Learning}. In \bibinfo{booktitle}{\emph{Conference on Fairness,
  Accountability and Transparency}}.
\newblock


\bibitem[\protect\citeauthoryear{Evgeniou and Pontil}{Evgeniou and
  Pontil}{2004}]%
        {evgeniou2004regularized}
\bibfield{author}{\bibinfo{person}{T. Evgeniou} {and} \bibinfo{person}{M.
  Pontil}.} \bibinfo{year}{2004}\natexlab{}.
\newblock \showarticletitle{Regularized multi--task learning}. In
  \bibinfo{booktitle}{\emph{ACM SIGKDD international conference on Knowledge
  discovery and data mining}}.
\newblock


\bibitem[\protect\citeauthoryear{Feldman, Friedler, Moeller, Scheidegger, and
  Venkatasubramanian}{Feldman et~al\mbox{.}}{2015}]%
        {feldman2015certifying}
\bibfield{author}{\bibinfo{person}{M. Feldman}, \bibinfo{person}{S.~A.
  Friedler}, \bibinfo{person}{J. Moeller}, \bibinfo{person}{C. Scheidegger},
  {and} \bibinfo{person}{S. Venkatasubramanian}.}
  \bibinfo{year}{2015}\natexlab{}.
\newblock \showarticletitle{Certifying and removing disparate impact}. In
  \bibinfo{booktitle}{\emph{International Conference on Knowledge Discovery and
  Data Mining}}.
\newblock


\bibitem[\protect\citeauthoryear{Hardt, Price, and Srebro}{Hardt
  et~al\mbox{.}}{2016}]%
        {hardt2016equality}
\bibfield{author}{\bibinfo{person}{M. Hardt}, \bibinfo{person}{E. Price}, {and}
  \bibinfo{person}{N. Srebro}.} \bibinfo{year}{2016}\natexlab{}.
\newblock \showarticletitle{Equality of opportunity in supervised learning}. In
  \bibinfo{booktitle}{\emph{Advances in neural information processing
  systems}}.
\newblock


\bibitem[\protect\citeauthoryear{{IBM}}{{IBM}}{2018}]%
        {cplex2018ibm}
\bibfield{author}{\bibinfo{person}{{IBM}}.} \bibinfo{year}{2018}\natexlab{}.
\newblock \bibinfo{title}{{User-Manual CPLEX 12.7.1}}.
\newblock \bibinfo{howpublished}{IBM Software Group}.
\newblock


\bibitem[\protect\citeauthoryear{Kamiran and Calders}{Kamiran and
  Calders}{2009}]%
        {kamiran2009classifying}
\bibfield{author}{\bibinfo{person}{F. Kamiran} {and} \bibinfo{person}{T.
  Calders}.} \bibinfo{year}{2009}\natexlab{}.
\newblock \showarticletitle{Classifying without discriminating}. In
  \bibinfo{booktitle}{\emph{International Conference on Computer, Control and
  Communication}}.
\newblock


\bibitem[\protect\citeauthoryear{Kamiran and Calders}{Kamiran and
  Calders}{2010}]%
        {kamiran2010classification}
\bibfield{author}{\bibinfo{person}{F. Kamiran} {and} \bibinfo{person}{T.
  Calders}.} \bibinfo{year}{2010}\natexlab{}.
\newblock \showarticletitle{Classification with no discrimination by
  preferential sampling}. In \bibinfo{booktitle}{\emph{Machine Learning
  Conference}}.
\newblock


\bibitem[\protect\citeauthoryear{Kamiran and Calders}{Kamiran and
  Calders}{2012}]%
        {kamiran2012data}
\bibfield{author}{\bibinfo{person}{F. Kamiran} {and} \bibinfo{person}{T.
  Calders}.} \bibinfo{year}{2012}\natexlab{}.
\newblock \showarticletitle{Data preprocessing techniques for classification
  without discrimination}.
\newblock \bibinfo{journal}{\emph{Knowledge and Information Systems}}
  \bibinfo{volume}{33}, \bibinfo{number}{1} (\bibinfo{year}{2012}),
  \bibinfo{pages}{1--33}.
\newblock


\bibitem[\protect\citeauthoryear{Kamishima, Akaho, and Sakuma}{Kamishima
  et~al\mbox{.}}{2011}]%
        {kamishima2011fairness}
\bibfield{author}{\bibinfo{person}{T. Kamishima}, \bibinfo{person}{S. Akaho},
  {and} \bibinfo{person}{J. Sakuma}.} \bibinfo{year}{2011}\natexlab{}.
\newblock \showarticletitle{Fairness-aware learning through regularization
  approach}. In \bibinfo{booktitle}{\emph{International Conference on Data
  Mining Workshops}}.
\newblock


\bibitem[\protect\citeauthoryear{Kearns, Neel, Roth, and Wu}{Kearns
  et~al\mbox{.}}{2017}]%
        {kearns2017preventing}
\bibfield{author}{\bibinfo{person}{M. Kearns}, \bibinfo{person}{S. Neel},
  \bibinfo{person}{A. Roth}, {and} \bibinfo{person}{Z.~S. Wu}.}
  \bibinfo{year}{2017}\natexlab{}.
\newblock \showarticletitle{Preventing Fairness Gerrymandering: Auditing and
  Learning for Subgroup Fairness}.
\newblock \bibinfo{journal}{\emph{arXiv preprint arXiv:1711.05144}}
  (\bibinfo{year}{2017}).
\newblock


\bibitem[\protect\citeauthoryear{Khosla, Zhou, Malisiewicz, Efros, and
  Torralba}{Khosla et~al\mbox{.}}{2012}]%
        {Efros}
\bibfield{author}{\bibinfo{person}{Aditya Khosla}, \bibinfo{person}{Tinghui
  Zhou}, \bibinfo{person}{Tomasz Malisiewicz}, \bibinfo{person}{Alexei~A
  Efros}, {and} \bibinfo{person}{Antonio Torralba}.}
  \bibinfo{year}{2012}\natexlab{}.
\newblock \showarticletitle{Undoing the damage of dataset bias}. In
  \bibinfo{booktitle}{\emph{European Conference on Computer Vision}}. Springer,
  \bibinfo{pages}{158--171}.
\newblock


\bibitem[\protect\citeauthoryear{Menon and Williamson}{Menon and
  Williamson}{2018}]%
        {menon2018cost}
\bibfield{author}{\bibinfo{person}{A.~K. Menon} {and} \bibinfo{person}{R.~C.
  Williamson}.} \bibinfo{year}{2018}\natexlab{}.
\newblock \showarticletitle{The cost of fairness in binary classification}. In
  \bibinfo{booktitle}{\emph{Conference on Fairness, Accountability and
  Transparency}}.
\newblock


\bibitem[\protect\citeauthoryear{Pedreshi, Ruggieri, and Turini}{Pedreshi
  et~al\mbox{.}}{2008}]%
        {pedreshi2008discrimination}
\bibfield{author}{\bibinfo{person}{D. Pedreshi}, \bibinfo{person}{S. Ruggieri},
  {and} \bibinfo{person}{F. Turini}.} \bibinfo{year}{2008}\natexlab{}.
\newblock \showarticletitle{Discrimination-aware data mining}. In
  \bibinfo{booktitle}{\emph{ACM SIGKDD international conference on Knowledge
  discovery and data mining}}.
\newblock


\bibitem[\protect\citeauthoryear{P{\'e}rez-Suay, Laparra, Mateo-Garc{\'\i}a,
  Mu{\~{n}}oz-Mar{\'\i}, G{\'o}mez-Chova, and Camps-Valls}{P{\'e}rez-Suay
  et~al\mbox{.}}{2017}]%
        {Prez-Suay2017Fair}
\bibfield{author}{\bibinfo{person}{A. P{\'e}rez-Suay}, \bibinfo{person}{V.
  Laparra}, \bibinfo{person}{G. Mateo-Garc{\'\i}a}, \bibinfo{person}{J.
  Mu{\~{n}}oz-Mar{\'\i}}, \bibinfo{person}{L. G{\'o}mez-Chova}, {and}
  \bibinfo{person}{G. Camps-Valls}.} \bibinfo{year}{2017}\natexlab{}.
\newblock \showarticletitle{Fair Kernel Learning}. In
  \bibinfo{booktitle}{\emph{Machine Learning and Knowledge Discovery in
  Databases}}.
\newblock


\bibitem[\protect\citeauthoryear{Pleiss, Raghavan, Wu, Kleinberg, and
  Weinberger}{Pleiss et~al\mbox{.}}{2017}]%
        {pleiss2017fairness}
\bibfield{author}{\bibinfo{person}{G. Pleiss}, \bibinfo{person}{M. Raghavan},
  \bibinfo{person}{F. Wu}, \bibinfo{person}{J. Kleinberg}, {and}
  \bibinfo{person}{K.~Q. Weinberger}.} \bibinfo{year}{2017}\natexlab{}.
\newblock \showarticletitle{On fairness and calibration}. In
  \bibinfo{booktitle}{\emph{Advances in Neural Information Processing
  Systems}}.
\newblock


\bibitem[\protect\citeauthoryear{Shalev-Shwartz and Ben-David}{Shalev-Shwartz
  and Ben-David}{2014}]%
        {shalev2014understanding}
\bibfield{author}{\bibinfo{person}{S. Shalev-Shwartz} {and} \bibinfo{person}{S.
  Ben-David}.} \bibinfo{year}{2014}\natexlab{}.
\newblock \bibinfo{booktitle}{\emph{Understanding machine learning: From theory
  to algorithms}}.
\newblock \bibinfo{publisher}{Cambridge University Press}.
\newblock


\bibitem[\protect\citeauthoryear{Shawe-Taylor and Cristianini}{Shawe-Taylor and
  Cristianini}{2004}]%
        {shawe2004kernel}
\bibfield{author}{\bibinfo{person}{J. Shawe-Taylor} {and} \bibinfo{person}{N.
  Cristianini}.} \bibinfo{year}{2004}\natexlab{}.
\newblock \bibinfo{booktitle}{\emph{Kernel methods for pattern analysis}}.
\newblock \bibinfo{publisher}{Cambridge University Press}.
\newblock


\bibitem[\protect\citeauthoryear{Smola and Sch{\"o}lkopf}{Smola and
  Sch{\"o}lkopf}{2001}]%
        {smola2001}
\bibfield{author}{\bibinfo{person}{A.~J. Smola} {and} \bibinfo{person}{B.
  Sch{\"o}lkopf}.} \bibinfo{year}{2001}\natexlab{}.
\newblock \bibinfo{booktitle}{\emph{Learning with Kernels}}.
\newblock \bibinfo{publisher}{MIT Press}.
\newblock


\bibitem[\protect\citeauthoryear{Woodworth, Gunasekar, Ohannessian, and
  Srebro}{Woodworth et~al\mbox{.}}{2017}]%
        {woodworth2017learning}
\bibfield{author}{\bibinfo{person}{B. Woodworth}, \bibinfo{person}{S.
  Gunasekar}, \bibinfo{person}{M.~I. Ohannessian}, {and} \bibinfo{person}{N.
  Srebro}.} \bibinfo{year}{2017}\natexlab{}.
\newblock \showarticletitle{Learning non-discriminatory predictors}. In
  \bibinfo{booktitle}{\emph{Computational Learning Theory}}.
\newblock


\bibitem[\protect\citeauthoryear{Zafar, Valera, Gomez~Rodriguez, and
  Gummadi}{Zafar et~al\mbox{.}}{2017a}]%
        {zafar2017fairness}
\bibfield{author}{\bibinfo{person}{M.~B. Zafar}, \bibinfo{person}{I. Valera},
  \bibinfo{person}{M. Gomez~Rodriguez}, {and} \bibinfo{person}{K.~P. Gummadi}.}
  \bibinfo{year}{2017}\natexlab{a}.
\newblock \showarticletitle{Fairness beyond disparate treatment \& disparate
  impact: Learning classification without disparate mistreatment}. In
  \bibinfo{booktitle}{\emph{International Conference on World Wide Web}}.
\newblock


\bibitem[\protect\citeauthoryear{Zafar, Valera, Gomez~Rodriguez, and
  Gummadi}{Zafar et~al\mbox{.}}{2017b}]%
        {zafar2017fairnessARXIV}
\bibfield{author}{\bibinfo{person}{M.~B. Zafar}, \bibinfo{person}{I. Valera},
  \bibinfo{person}{M. Gomez~Rodriguez}, {and} \bibinfo{person}{K.~P. Gummadi}.}
  \bibinfo{year}{2017}\natexlab{b}.
\newblock \showarticletitle{Fairness constraints: Mechanisms for fair
  classification}. In \bibinfo{booktitle}{\emph{International Conference on
  Artificial Intelligence and Statistics}}.
\newblock


\bibitem[\protect\citeauthoryear{Zafar, Valera, Rodriguez, Gummadi, and
  Weller}{Zafar et~al\mbox{.}}{2017c}]%
        {zafar2017parity}
\bibfield{author}{\bibinfo{person}{M.~B. Zafar}, \bibinfo{person}{I. Valera},
  \bibinfo{person}{M. Rodriguez}, \bibinfo{person}{K. Gummadi}, {and}
  \bibinfo{person}{A. Weller}.} \bibinfo{year}{2017}\natexlab{c}.
\newblock \showarticletitle{From parity to preference-based notions of fairness
  in classification}. In \bibinfo{booktitle}{\emph{Advances in Neural
  Information Processing Systems}}.
\newblock


\bibitem[\protect\citeauthoryear{Zemel, Wu, Swersky, Pitassi, and Dwork}{Zemel
  et~al\mbox{.}}{2013}]%
        {zemel2013learning}
\bibfield{author}{\bibinfo{person}{R. Zemel}, \bibinfo{person}{Y. Wu},
  \bibinfo{person}{K. Swersky}, \bibinfo{person}{T. Pitassi}, {and}
  \bibinfo{person}{C. Dwork}.} \bibinfo{year}{2013}\natexlab{}.
\newblock \showarticletitle{Learning fair representations}. In
  \bibinfo{booktitle}{\emph{International Conference on Machine Learning}}.
\newblock


\end{thebibliography}
\end{document}